\documentclass{article}

\usepackage[preprint]{corl_2026} 

\usepackage[utf8]{inputenc} 
\usepackage[T1]{fontenc}    
\usepackage{hyperref}       
\usepackage{url}            
\usepackage{booktabs}       
\usepackage{amsfonts}       
\usepackage{nicefrac}       
\usepackage{microtype}      
\usepackage{xcolor}         

\usepackage{graphicx}
\usepackage{subcaption}
\usepackage{algorithm}
\usepackage[noEnd,commentColor=black]{algpseudocodex}
\usepackage[most]{tcolorbox}
\usepackage{float}
\usepackage{wrapfig}
\usepackage{lipsum}
\usepackage{colortbl}
\usepackage{soul}
\usepackage{enumitem}
\usepackage{multirow}
\usepackage{makecell}
\usepackage{tablefootnote}
\usepackage[para,online,flushleft]{threeparttable}
\usepackage[nameinlink,noabbrev]{cleveref}
\usepackage{siunitx}
\usepackage{amsmath}
\usepackage{wrapfig} 

\definecolor{myblue}{rgb}{1,0.5,0}
\definecolor{urlteal}{HTML}{0F766E}
\hypersetup{
    colorlinks = true,
    linkcolor = {orange},
    citecolor = {orange},
    urlcolor=orange,
    anchorcolor = black,
}

\newcommand{\ours}[1]{\textcolor{black}{EXPO-FT}{#1}}

\title{EXPO-FT: Sample-Efficient Reinforcement Learning Finetuning for Vision-Language-Action Models}

\newcommand{\symfootnotetext}[2]{%
  \begingroup
  \renewcommand{\thefootnote}{#1}%
  \footnotetext{#2}%
  \endgroup
}

\author{
 Perry Dong$^{*, \dagger}$ \quad Kuo-Han Hung$^{*}$ \quad Tian Gao \quad Dorsa Sadigh \quad Chelsea Finn \\[2pt]
  Stanford University \\[2pt]
  \href{https://pd-perry.github.io/expo-ft}{\texttt{https://pd-perry.github.io/expo-ft/}}
}

\begin{document}
\maketitle
\symfootnotetext{*}{Equal contributions. }
\symfootnotetext{$\dagger$}{Corresponding author: \texttt{perryd@stanford.edu}}
\setcounter{footnote}{0}


\begin{abstract}
    The ability to efficiently and reliably learn new tasks has been a foundational challenge in robotics. Vision-Language-Action (VLA) models have demonstrated strong generalization across diverse manipulation tasks, yet pretrained policies consistently fall short of the reliability required for real-world deployment. Reinforcement learning (RL) fine-tuning offers a promising path to bridge this gap, but existing approaches either train from scratch without fully leveraging pretrained priors, or fine-tune VLAs without achieving the sample efficiency and success rates that practical deployment demands. We present \ours{}, a system for stable, sample-efficient RL finetuning of pretrained VLA policies that closes this gap. Our system solves a suite of challenging manipulation tasks, including routing string lights and inserting the plug to light it up, striking a pool ball into a pocket, and inserting a flower into a wine bottle, each requiring combinations of high precision, dynamic actions, and robustness to varied initial states. Our system achieves perfect task performance (30/30 successes) across all evaluated tasks within an average of 19.1 minutes of online robot data, outperforming both prior RL-from-scratch and VLA finetuning approaches. We release an open-source codebase with the aim of facilitating broader adoption of RL finetuning of VLA models in robotics.

\end{abstract}

\section{Introduction} \label{sec:introduction}

The ability to efficiently and reliably learn new tasks has been a foundational challenge in robotics. Recent advances in large-scale imitation learning with Vision-Language-Action models (VLAs) have yielded pretrained policies capable of executing complex, multi-step behaviors across diverse task settings ~\citep{intelligence2025pi05visionlanguageactionmodelopenworld, geminiroboticsteam2025geminirobotics15pushing}. Yet despite their promise, even the most capable pretrained models fall persistently short of achieving reliable task success for deployment, which is consequential as failures in the real world are costly. In this work, we study the problem of finetuning VLAs with RL efficiently and stably to a reliable performance. We believe bridging this gap can unlock a new class of policies capable of reliable, real-world deployment.

Existing approaches of training real-world robot policies via RL fall into two camps. The first comprises methods that train RL policies without finetuning a VLA ~\citep{luo2025precisedexterousroboticmanipulation,luo2025serlsoftwaresuitesampleefficient,xu2026rltokenbootstrappingonline,lei2026rl100performantroboticmanipulation}, either from scratch or training lightweight models on top of the VLA. While some of these approaches achieve strong asymptotic performance, including near-perfect success rates in certain settings~\citep{luo2025precisedexterousroboticmanipulation}, they cannot be used to finetune pretrained VLAs and often cannot leverage the rich semantic and behavioral priors embedded in large pretrained VLAs. The second category consists of methods that finetune pretrained VLAs, thereby inheriting their generalizable representations~\citep{ren2024diffusionpolicypolicyoptimization,wagenmaker2025steeringdiffusionpolicylatent,li2025grrlgoingdexterousprecise,li2025simplevlarlscalingvlatraining,chen2026pitextttrlonlinerlfinetuning}. However, these methods either fail to reach a reliable success rate, require prohibitively large numbers of environment interactions, or both. The gap of frameworks that simultaneously exploits the prior of a pretrained VLA and achieves the sample efficiency and reliability necessary for practical deployment motivates our work.

In this paper, we develop a system for finetuning pretrained robotic policies with reinforcement learning. Our system, \ours{}, demonstrates that RL finetuning of VLA models can be both sample efficient and reliable. We elucidate design choices that address specific challenges of reinforcement learning on top of pretrained models in robotics. To address the challenge of sample efficient finetuning of pretrained robotics policies, we build on the recently proposed EXPO algorithm~\citep{dong2026expostablereinforcementlearning}, which provides a principled foundation for RL fine-tuning in this regime. We extend EXPO to incorporate temporally extended actions that most modern VLAs operate over. We  also integrate human-in-the-loop feedback~\citep{kelly2019hgdaggerinteractiveimitationlearning} to provide corrections during online training, enabling targeted interventions that improve sample efficiency where autonomous exploration is insufficient.
As a concrete instantiation of our framework, we adopt $\pi_{0.5}$~\citep{intelligence2025pi05visionlanguageactionmodelopenworld} as the backbone VLA model—a state-of-the-art generalist policy that has demonstrated strong performance across a diverse range of manipulation tasks. 

\begin{wrapfigure}{r}{0.5\textwidth}
  \vspace{-10pt} 
  \centering
  \includegraphics[width=0.48\textwidth]{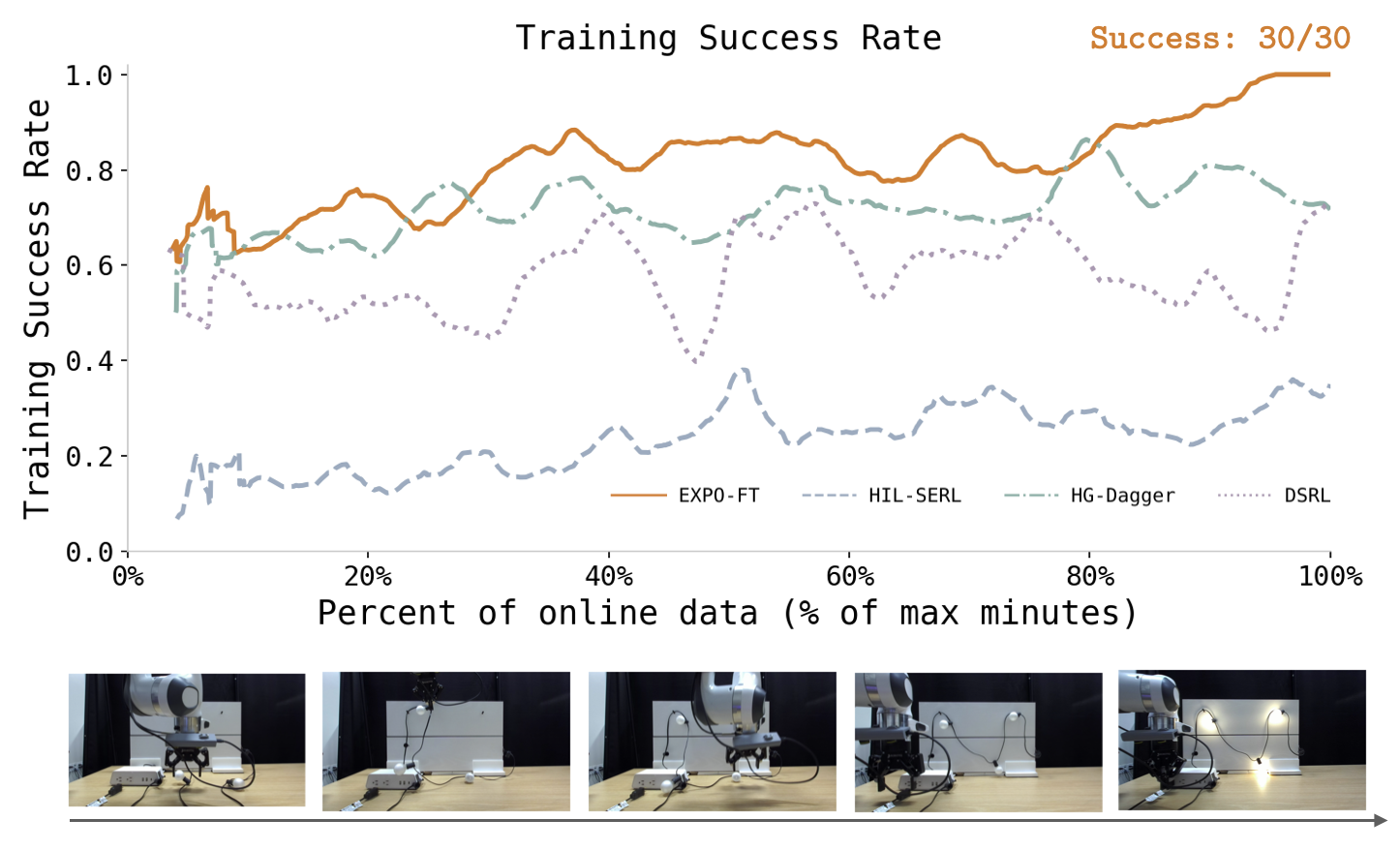}
  \caption{\footnotesize \textbf{Average training success rates} of \ours{} compared to prior methods. \ours{} achieves a reliable performance with high sample efficiency where prior methods often do not converge reliably. }
  \label{fig:fig1}
  \vspace{-10pt} 
\end{wrapfigure}

We empirically find that our system achieves dexterous and precise manipulation capabilities across a diverse set of challenging tasks, including routing string lights and inserting the power connector to illuminate them, striking a pool ball into a pocket, and inserting a flower into a wine bottle. These tasks require varying combinations of precision, adaptability to diverse initial states, and dynamic actions in the case of striking a pool ball, collectively representing significant challenges for robotic manipulation. Compared to state-of-the-art prior methods, our system outperforms both methods that perform reinforcement learning from scratch and approaches that build on pretrained models, as shown in~\Cref{fig:fig1}. Notably, the full combination of our proposed components achieves the high reliability that is necessary for real deployment scenarios (30/30 success) across all of these complex tasks in an average of 20 minutes of online data.

Our primary contributions are a system for highly sample-efficient, reliable finetuning of VLA models with reinforcement learning, and an easy-to-use, open-source codebase to support such fine-tuning. We demonstrate that, with the right design, RL finetuning of VLAs can be sample efficient and achieve highly reliable success rates. 
We hope this work serves as a stepping stone toward broader adoption of finetuning generalist VLA models with reinforcement learning in the robotics community.

\section{Related Work} \label{sec:related_work}

\textbf{Real-world reinforcement learning.}
Reinforcement learning has a long history of real-world deployment in robotics~\citep{haarnoja2018soft,kalashnikov2018scalable,JMLR:v17:15-522,6095096,deisenroth,Kietzmann2009TheNS,5509672,PETERS2008682}. In real-world reinforcement learning, sample efficiency remains a central challenge addressed by numerous prior algorithmic works~\citep{dong2026expostablereinforcementlearning,ball2023efficientonlinereinforcementlearning,dong2026reallyneedpretrainqfunctions,dong2025reinforcementlearningimplicitimitation,chen2021randomizedensembleddoubleqlearning, nauman2024biggerregularizedoptimisticscaling}. Recent efforts have explored assembling these components together into sample-efficient real-world RL systems~\citep{luo2025precisedexterousroboticmanipulation,luo2025serlsoftwaresuitesampleefficient, lei2026rl100performantroboticmanipulation,ankile2025residualoffpolicyrlfinetuning,ICLR2024_9c537882,dong2025mattersbatchonlinereinforcement}, with extensions such as human-in-the-loop feedback~\citep{luo2025precisedexterousroboticmanipulation} providing further gains in sample efficiency. While many of these systems achieve high task success rates, they rely on small Gaussian policies that cannot leverage large pretrained priors. In contrast to prior work, we approach real-world RL through the lens of finetuning a pretrained VLA model, leveraging it simultaneously as a generalist policy capable of complex manipulation and as an informative prior that enables sample-efficient online learning.

\textbf{Reinforcement Learning with VLAs.}
There has been growing interest in finetuning VLAs with RL. Classical RL algorithms for continuous control are designed around Gaussian policies, which are not directly applicable to many modern VLAs that use more expressive policy classes such as flow matching or diffusion. Several works address this mismatch by developing RL algorithms tailored to flow or diffusion policies~\citep{dong2026expostablereinforcementlearning,wagenmaker2025steeringdiffusionpolicylatent,yang2023policyrepresentationdiffusionprobability,ren2024diffusionpolicypolicyoptimization,mark2024policyagnosticrloffline,dong2026fastervalueguidedsamplingfast,psenka2025learningdiffusionmodelpolicy,li2026qlearningadjointmatching}. $\pi_{0.6}$~\citep{intelligence2025pi06vlalearnsexperience} trains a VLA policy with advantage conditioning but in an offline setting whereas we focus on online finetuning, allowing for more data efficient learning. Multiple online RL methods use on-policy algorithms to finetune the VLA~\citep{liu2026rlbringvlageneralization,li2025simplevlarlscalingvlatraining,tan2025interactiveposttrainingvisionlanguageactionmodels,lu2025vlarlmasterfulgeneralrobotic,ren2024diffusionpolicypolicyoptimization,chen2026pitextttrlonlinerlfinetuning}. We instead use off-policy RL for increased sample efficiency. Existing methods for using off-policy RL with VLAs often either train a lightweight auxiliary policy~\citep{xu2026rltokenbootstrappingonline,chen2025conrftreinforcedfinetuningmethod, guo2025improvingvisionlanguageactionmodelonline,xiao2025selfimprovingvisionlanguageactionmodelsdata,yuan2024policydecoratormodelagnosticonline}  or perform optimization in a latent space~\citep{wagenmaker2025steeringdiffusionpolicylatent,li2025grrlgoingdexterousprecise,zhang2025alignthensteeradaptingvisionlanguageaction}. Our work differs from these prior methods along a few key dimensions. First, rather than only training a separate lightweight policy or decoupling VLA training from the RL loop, we directly finetune the full VLA within a unified RL pipeline, yielding more coherent training. Second, we incorporate human-in-the-loop interventions to accelerate learning. Third, many prior methods~\citep{chen2025conrftreinforcedfinetuningmethod,xiao2025selfimprovingvisionlanguageactionmodelsdata,yuan2024policydecoratormodelagnosticonline} operate on single-step actions. In contrast, we optimize over the action chunks natively produced by the VLA, preserving the temporal abstraction central to its design. Together, these design choices yield an efficient and performant VLA finetuning framework, as demonstrated by our experimental results.

\section{Background} \label{sec:background}
We consider the standard reinforcement learning framework, where problems are modeled by a Markov decision process (MDP) $\mathcal{M} = (\mathcal{S}, \mathcal{A}, r, T, \gamma, \rho)$. In the MDP, $\mathcal{S}$ is the state space and $\mathcal{A}$ is the action space. At each timestep $t$, the agent observes state $s_t \in \mathcal{S}$ and selects action $a_t \in \mathcal{A}$ according to policy $\pi : \mathcal{S} \rightarrow \mathcal{A}$, and receives scalar reward $r(s_t, a_t) \in \mathbb{R}$. The environment transitions following the transition dynamics of the MDP $s_{t+1} \sim T(\cdot \mid s_t, a_t)$, with starting state initialized from $s_0 \sim \rho(\cdot)$. The tuple $(s_t, a_t, r_t, s_{t+1})$ is added in the replay buffer $\mathcal{D}$ for learning. The goal of reinforcement learning is to maximizes the expected discounted return $\mathbb{E}_{\pi}\!\left[\sum_{t=0}^{T} \gamma^t \, r(s_t, a_t)\right]$, where $\gamma \in [0, 1]$ is the discount factor.

\textbf{EXPO~\citep{dong2026expostablereinforcementlearning}. } To finetune the VLA policy with RL and leverage the prior of the pretraining model, we build on the recently proposed highly sample efficient and stable RL algorithm for training expressive flow or diffusion policies, EXPO~\citep{dong2026expostablereinforcementlearning}. Unlike classical sample-efficient RL algorithms which are designed around Gaussian policies and cannot be applied to pretrained VLAs that use flow or diffusion policies, EXPO provides a principled foundation for RL fine-tuning in this regime. 

EXPO maintains two parameterized policies: a base flow policy, which in our case is the VLA model $\pi_\text{VLA}$ trained with a supervised loss, and a lightweight edit policy $\pi_\text{edit}$ trained to maximize the Q-function:
{\small
\begin{equation}
\begin{split}
    \mathcal{L}(\pi_\text{edit}) = -\mathbb{E}_{(s_t,a_{t})\sim\mathcal{D},\;\hat{a}_{t}\sim\pi_\text{edit}}
    \bigl[Q_\phi(s_t,\, a_{t} + \hat{a}_{t}) - \alpha \log \pi_\text{edit}(\hat{a}_{t}|s_t, a_{t})\bigr]
\end{split}
\end{equation}
}

The edit policy predicts an edit $\hat{a}$ constrained to $[-\beta, \beta]$ that is added to the base action $a$ to produce the edited actions $\tilde{a} = a + \hat{a}$. This avoids gradient backpropagation through the VLA model while still grounding TD updates in near-optimal actions. The final inference and TD backup policy is an on-the-fly (OTF) policy that selects the value-maximizing candidate across base and edited actions:
\begin{equation}
    \tilde{a}^* = \underset{a \;\in\; \bigcup_{i=1}^{N}\{a_i,\,\tilde{a}_i\}}{\arg\max}\; Q_\phi(s, a)
\end{equation}

with the Q-function trained via TD learning:
\begin{equation}
    \mathcal{L}(\phi) = \mathbb{E}_{(s_t, a_{t}, s_{t+1}) \sim \mathcal{D}}\left[\left(r_t + \gamma Q_{\phi'}(s_{t+1}, \tilde{a}^*_{t+1}) - Q_\phi(s_t, a_{t})\right)^2\right].
\end{equation}

\section{EXPO-FT} \label{sec:method}

\begin{figure}[t]
  \centering
  \includegraphics[width=\linewidth]{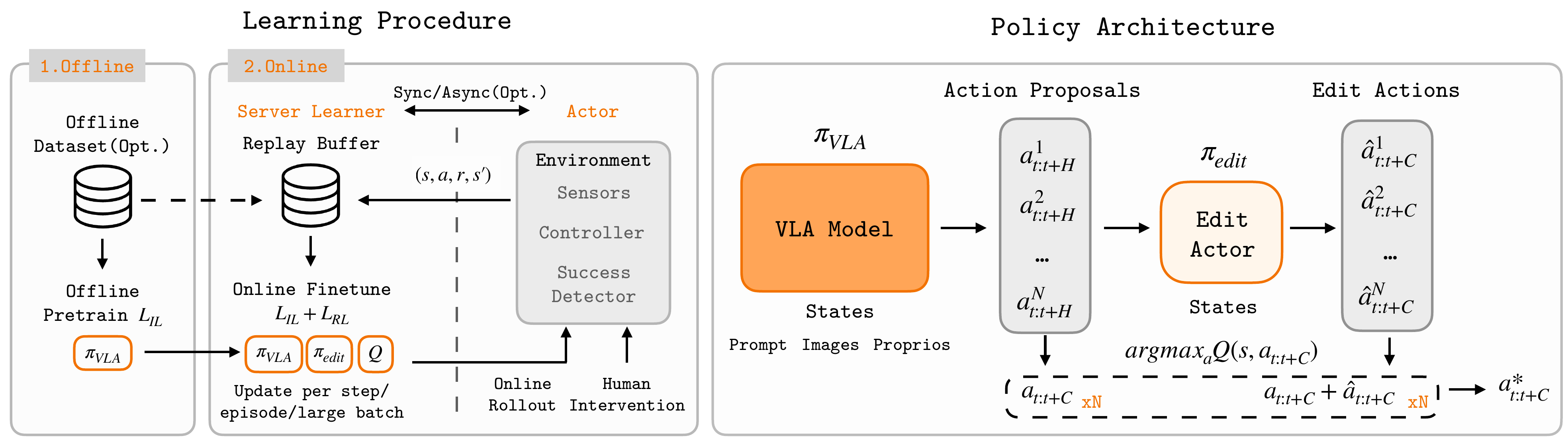}
  \caption{\footnotesize \textbf{Left: Overview of \ours{}.} \ours{} features a server that handles VLA training and inference and a learner process that steps in the environment to enable VLA finetuning with RL. \textbf{Right: Architecture of \ours{}}. \ours{} finetunes the VLA model with EXPO for sample-efficient training. } 
  \label{fig:method}
\end{figure}

In this section, we present our complete system for sample-efficient, reliable finetuning for VLA models with reinforcement learning. Our central design goal is to finetune VLA models to a performance necessary for real-world deployment with high sample efficiency across different use cases. We start by describing the problem statement (\Cref{subsec:problem_statement}), then describe the approach used for finetuning (\Cref{subsec:alg}), and lastly describe the implementation details (\Cref{subsec:design}). The complete system of \ours{} is summarized in~\Cref{fig:method}.

\subsection{Problem Statement}\label{subsec:problem_statement}

We consider the problem of finetuning a pretrained vision-language-action (VLA) model $\pi_\text{VLA}$, which often is a generalist policy trained on a diverse corpus of demonstrations and conditioned on natural language instructions to specify the desired task, with reinforcement learning. Modern VLAs often employs action chunking, predicting a sequence of
$H$ future actions $a_{t:t+H}$ at each timestep. Because of the rich behavioral prior acquired during pretraining, VLA policies are often capable of completing tasks zero-shot to some success rate. However, for tasks that $\pi_\text{VLA}$ does not reach a reasonable zero-shot performance, we follow standard offline-to-online RL practices and assume access to a small offline dataset of expert demonstrations $\mathcal{D}_0$, collected via human teleoperation,, which is used to bootstrap the finetuning process. The dataset is collected by a human teleoperation. We adopt a sparse binary reward $r \in \lbrack0,1\rbrack$ indicating successful task completion. The task completion classifier may be either rule-based or learned. Observations consist of multi-view RGB images—comprising a wrist-mounted camera and a fixed side-view camera—augmented with robot proprioceptive state. Policy parameters are updated either after every environment step, at the end of each episode, or at fixed episode-batch intervals. The objective is to maximize task success rate.

A central requirement of our setting is practicality on real robots, which places strong constraints on sample efficiency and algorithmic compatibility. To this end, we build upon EXPO because of its sample efficiency and compatibility with diffusion- and flow-matching-based policy architectures, which are predominant in modern VLAs. However, the original formulation of EXPO neither supports action chunking nor incorporates human-in-the-loop feedback. The former is important because most modern VLAs operate by predicting multi-step action chunks rather than single actions; the latter has been shown to significantly improve sample efficiency. In the following section, we discuss how these challenges are addressed in \ours{}.

\subsection{Sample-efficient VLA finetuning with RL}\label{subsec:alg}

\textbf{Temporally extended actions. } Rather than predicting single-step actions, most modern VLAs operate over temporally extended action sequences, outputting a chunk of $H$ consecutive actions $a_{t:t+H}=(a_t, a_{t+1}, \ldots, a_{t+H-1})$ and executing $C\leq H$ at each timestep. On top of EXPO, we incorporate temporally extended actions to be compatible with the native interface of most modern VLAs. We adapt the edit policy and Q-function to operate over chunks of executed actions $a_{t:t+C}$. Concretely, the edit policy predicts edits $\hat a_{t:t+C}$ and the edited actions that are selected are $\tilde a^i_{t:t+C} = a^i_{t:t+C} + \hat a^i_{t:t+C}$. The edit policy loss is 
{\small
\begin{equation}
\begin{split}
    \mathcal{L}(\pi_\text{edit}) = -\mathbb{E}_{(s_t,a_{t:t+C})\sim\mathcal{D},\;\hat{a}_{t:t+C}\sim\pi_\text{edit}}
    \bigl[Q_\phi(s_t,\, a_{t:t+C} + \hat{a}_{t:t+C}) - \alpha \log \pi_\text{edit}(\hat{a}_{t:t+C}|s_t, a_{t:t+C})\bigr]
\end{split}
\end{equation}
}

with the Q-function trained via TD learning:
\begin{equation}
    \mathcal{L}(\phi) = \mathbb{E}_{(s_t, a_{t:t+C}, s_{t+C}) \sim \mathcal{D}}\left[\left(r_t + \gamma Q_{\phi'}(s_{t+C}, \tilde{a}^*_{t+C:t+2\times C}) - Q_\phi(s_t, a_{t:t+C})\right)^2\right]
\end{equation}

The base VLA is updated using its original training objective, without modification.

\textbf{Human-in-the-loop interventions}. To make learning more sample-efficient and reliable, we use human-in-the-loop interventions during the training process. Specifically, during RL training, a human operator can intervene to provide action corrections when necessary. Interventions can be done for any individual actions within an action chunk. At any consecutive timesteps $t'$ to $t''$ within the action chunk $a_{t:t+C}$, the human can intervene with actions $\bar a_{t':t''}$ where the robot then performs the actions the human provided, and the corresponding timesteps are overwritten with the corrective action, keeping the rest of the action chunk the same yielding action sequence $(a_t, \ldots, \bar a_{t'},\ldots,\bar a_{t''} \ldots, a_{t+C-1})$. Interventions can happen across action chunks as well. The chunks of actions are then added to the replay buffer after execution. We find the interventions are especially helpful to provide correct signal to reduce the amount of exploration necessary.

\textbf{Training process. } 
We now describe the end-to-end training procedure of \ours{}. For each task, we begin by configuring the camera setup and defining the reward signal, which can take the form of a rule-based criterion, a learned binary classifier, or any alternative reward specification such as a learned reward model.
Before online RL begins, we assess whether the VLA can perform the task zero-shot at a reasonable success rate. If not, we collect a set of human demonstrations and finetune the VLA with imitation learning until it reaches a success rate around 40\% or above. This data can either be retained as a separate offline dataset or used to initialize the replay buffer. We initialized the replay buffer directly in our experiments as our dataset sizes are small. 
With the supervised finetuned VLA policy, we start online RL training. The actor executes rollouts in the environment, with a human operator available to intervene via teleoperation. The updates can be done per step, per episode, or per batch of episodes, depending on the task requirements and available compute.

\subsection{Implementation Details}
\label{subsec:design}

We make several design choices to allow flexibility of our system depending on the task and compute available. We first present these details of our system then describe the training process. 

\textbf{Vision backbones. } Visual inputs appear in both the actor and the critic from the side and wrist cameras. Since we use a VLA as the actor, it comes equipped with a pretrained visual encoder. For the critic, one natural option is to share this encoder, leveraging the same rich visual representations. However, the VLA encoder is large, making shared inference computationally prohibitive during the tight actor-learner loop. We find that equipping the critic with a separate, lightweight ResNet-50~\citep{he2015deepresiduallearningimage} encoder can achieve high task performance at substantially lower computational cost. We include full details on architecture choices in Appendix \Cref{appendix:train_setting}.

\textbf{Human interventions.} In our experiments, the human provides more interventions earlier in the training process where the robot need more corrections, and the intervention rate reduces to zero as the robot learns its policy to successful complete the task. We use a SpaceMouse to provide interventions. The human operator observes the policy actions and engages with the SpaceMouse when interventions are required. The actions from the SpaceMouse then automatically overrides the policy actions and gets executed by the robot.

\textbf{Learning procedure.} Training a VLA in an online RL loop presents a practical latency challenge as the model size is large and because of that, inference, environment interaction, and gradient updates all take a significant amount of time. We therefore decouple the system into a learner and an actor. The learner is responsible for VLA training and inference, and maintaining the data and replay buffer. The actor runs separately and communicates with the real-world environment to execute actions, collect human interventions, and send $(s,a,r,s')$ tuples back to the learner's replay buffer. The learner then updates the agent either per step, per episode, or per large batch of data. The first option is the most sample efficient given enough compute, whereas updating per episode or large batch of data can better satisfy the safety requirement of deployment. The learner can communicate both synchronously and asynchronously between the server and actor; the asynchronous mode allows gradient updates and environment interaction to proceed in parallel, improving overall training throughput when additional GPUs are available. We use synchronous communication as it is faster when GPU is limited (2 or under). This learner actor interface described is released with our open-source codebase. 

\textbf{Reward design.} An accurate reward function is important for obtaining a reliable performance for deployment. We adopt a sparse binary reward signal to avoid the burden of task-specific reward shaping. For each task, we specify a rule-based binary classifier that gives a positive reward only upon task completion and zero otherwise. For example, in the candy scoop task, we first verify that candies are present in the scoop once the scoop is raised above a height threshold above the bin, with candies determined by pixel color matching, then check whether the candy count decreases by at least a threshold amount within a specified xyz region corresponding to the target container. We defer the reward classifier of all tasks to Appendix \Cref{appendix:task_setting}. This approach is straightforward to set up, generalizes across diverse tasks, and sidesteps the engineering overhead of dense reward specification.

\section{Experiments} \label{sec:experiment}

\begin{figure}[t]
  \centering
  \includegraphics[width=\linewidth]{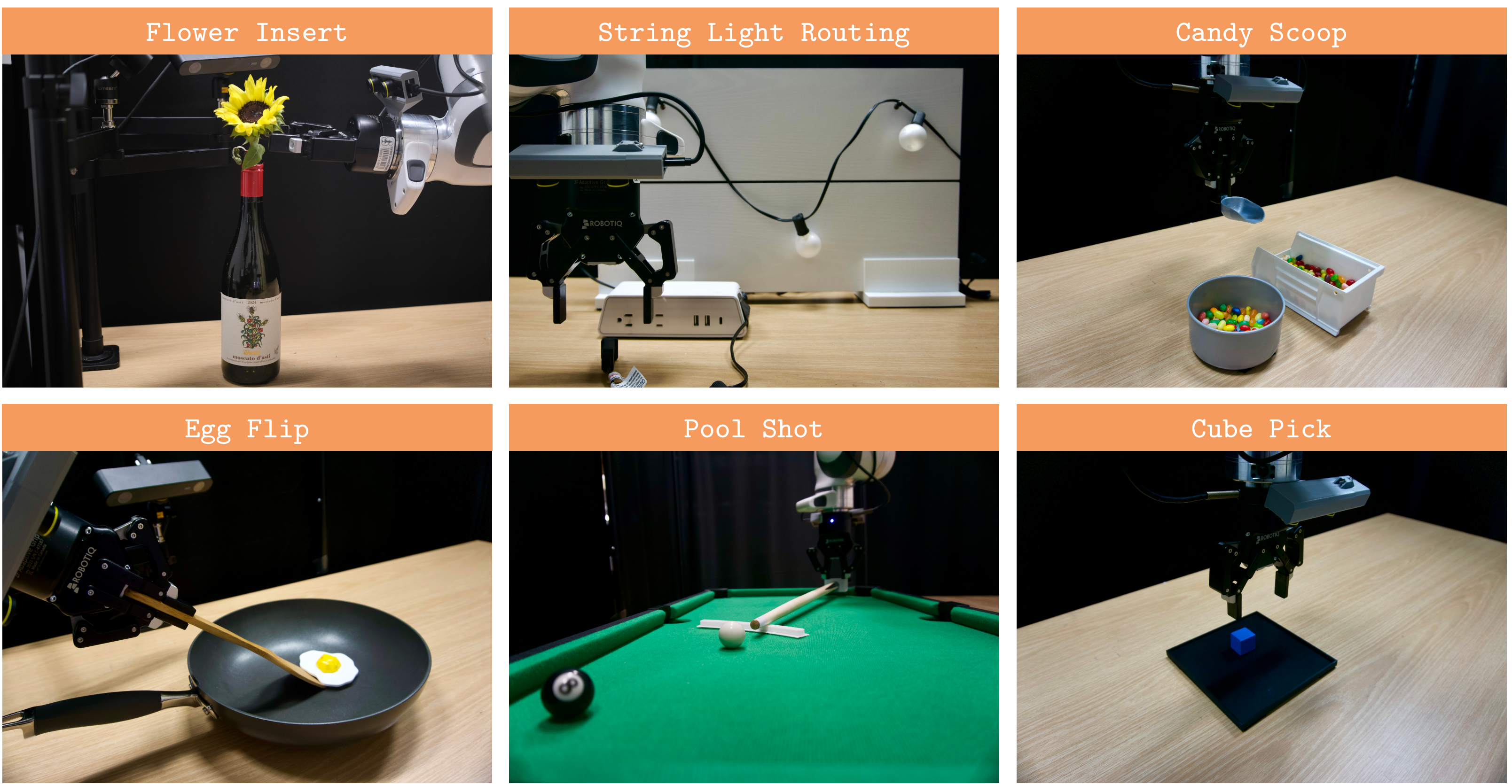}
  \caption{ \footnotesize \textbf{Eight real-world manipulation tasks in our evaluation suite.} Flower Insert (tight insertion tolerances), String Light Routing - RouteI/II, Insert (long-horizon precise alignment), Egg Flip (dynamic contact-rich tool use), Candy Scoop (stable control in visually messy scenes), Pool Shot (precise speed control) and Cube Pick (large scene randomization). The tasks span dexterous, precise, deformable, and dynamic manipulation.}
  \label{fig:tasks}
\end{figure}

We evaluate \ours{} on 8 challenging real-world robotic manipulation tasks requiring a combination of dynamic actions, high precision, and diverse initial states, and compare against strong prior methods for finetuning pretrained models and training from scratch to answer whether \ours{} can efficiently finetune VLAs with RL to a high performance and reliability. We start by presenting the experiment details and task setup, followed by the results.

\subsection{Experiment details } 
For all experiments, the robot is controlled using an end-effector action space consisting of Cartesian velocity commands and gripper velocity commands running at 10 HZ. The observations consist of two RGB images from side and wrist view cameras at 224x224 resolution and proprioceptive state representation, which includes the robot end-effector position and orientation. Environment resets are performed either automatically or manually, depending on the task.

We use a binary sparse reward indicating task completion for all tasks, with rule-based success detectors that achieve over 95\% accuracy in identifying successful task completion. For evaluation, a human observes if the task is successful. For each task, we report the success rate across 30 evaluation trials and the total amount of online data used for learning. During evaluation, initial states are randomized using either scripted robot motions or human resets. Detailed reward specifications, task success detectors, reset mechanisms, and task randomization strategies for each task are provided in Appendix~\Cref{appendix:task_setting}. Our experiments are performed on two NVIDIA H200 GPUs for 8k to 20k environment steps depending on how many steps it takes to converge for the task.

\subsection{Description of tasks } We select tasks that cover a broad range of manipulation challenges, including precise object alignment, flexible object handling, and dynamic interaction. An illustration of each task can be found in ~\Cref{fig:tasks} and ~\Cref{fig:task_stips}.

\textbf{Egg Flip. } In this task, the robot is required to flip a 3D printed fried egg in a pan using a spatula. The robot must slide the spatula underneath the egg, lift it with appropriate force and orientation, and execute a flipping motion without dropping or damaging the egg. The task is considered successful if the egg is fully flipped and remains inside the pan after manipulation. This task requires contact-rich manipulation and precise control of the spatula pose during dynamic interaction, where excessive force or inaccurate timing can cause the egg to slide, rotate unexpectedly, or fail to flip properly.

\textbf{String Light Routing - Route I \& II \& Insert. } The string light routing task requires the robot to complete a three-stage light assembly and activation process. The task is divided into three subtasks: hanging the first light bulb onto a nail, hanging the second light bulb onto another nail, and plugging the power cable into the power source to activate the lights. Each subtask requires precise alignment and manipulation within a limited workspace, where small positioning errors can prevent successful hanging or cable insertion. The success of each subtask is evaluated individually based on whether the corresponding light bulb is correctly hung or whether the power cable is successfully plugged in. This task requires accurate object alignment and stable control across sequential manipulation steps.

\textbf{Candy Scoop. } In this task, the robot is required to scoop candies from a container using a spoon and transfer them into another container. The robot must carefully control the spoon’s pose and motion to collect and transport the candies with minimal spillage. The task is considered successful if most of the candies are transferred into the target container. This task requires stable motion control and precise scooping depth selection, since shallow or overly deep scoops reduce success. It also requires robust handling of deformable and noisy candy distributions with varying colors, which increases perceptual and manipulation difficulty.

\textbf{Cube Pick. } In this task, the robot is required to pick up a cube from a surface and lift it. The robot must move its gripper to the cube, align properly, and grasp it securely without slipping. The task is considered successful if the cube is successfully lifted off the surface. This task requires accurate positioning and stable grasping under large scene randomization, including edge cases where cubes are placed near the edge of the plate, which increases the risk of failed grasps or unstable contacts.

\textbf{Flower Insert. } In this task, the robot is required to insert a flower stem into a corresponding vase or holder. The robot must align the stem with the narrow opening and carefully guide it into the slot with controlled downward motion. The task is considered successful if the flower is fully inserted. This task requires precise alignment under tight tolerances, where small position or orientation errors can lead to failed insertion or deformation of the stem.

\textbf{Pool Shot. } In this task, the robot is required to strike the cue ball to pocket the black ball into a designated hole. The robot must control the striking force applied to the cue ball while ensuring a clean hit. The task is considered successful if the black ball is successfully pocketed and the cue ball remains out of the hole. This task requires precise force control, where insufficient force fails to pocket the ball and excessive force leads to undesired cue ball trajectories.

\subsection{Comparisons}
We compare our approach to two categories of methods: (1) methods that train reinforcement learning policies from scratch in the real world, and (2) methods that finetune pretrained vision-language-action (VLA) models. We describe each method below.

\textbf{HIL-SERL~\citep{luo2025precisedexterousroboticmanipulation}.} HIL-SERL is a sample efficient real-world reinforcement learning system that incorporates human-in-the-loop interventions for policy learning, demonstrating reliable performance on complex manipulation tasks such as motherboard and IKEA furniture assembly. While HIL-SERL shares our design of leveraging human feedback to guide online learning, it is designed to train policies from scratch and cannot finetune a pretrained VLA, nor exploit the behavioral priors encoded within one. As a result, despite achieving highly reliable performance in its original evaluations, it struggles in more complex settings such as those with large initial-state distributions, where \ours{} can naturally handle both with a more sample efficient RL algorithm and by bootstrapping from a pretrained VLA.

\textbf{DSRL~\citep{wagenmaker2025steeringdiffusionpolicylatent}.} DSRL is a state-of-the-art reinforcement learning method for finetuning pretrained diffusion and flow-matching policies. By steering the latent noise, it learns the optimal actions within the prior, yielding substantial performance improvements over the base diffusion or flow policy. DSRL can be directly applied to VLA finetuning by perturbing the model's noise samples to improve downstream task performance. However, DSRL is constrained to optimizing within the modes of the prior distribution, which heavily limits its ability to acquire new behavior, and cannot incorporate human interventions during online execution. 

\textbf{HG-DAgger~\citep{kelly2019hgdaggerinteractiveimitationlearning}.} HG-DAgger is a human-gated variant of the DAgger~\citep{ross2011reductionimitationlearningstructured} algorithm that intermittently solicits expert corrections during online execution. HG-DAgger is a natural choice for improving VLA models because of its simplicity and direct applicability to pretrained VLAs; however, HG-DAgger relies exclusively on behavioral cloning from human interventions, optimizes no reward signal, and cannot leverage suboptimal experience collected during autonomous execution.

\textbf{Supervised finetuning.} As an additional reference point, we report the performance of standard supervised finetuning of the base VLA on the task demonstration data, prior to any online interaction.

\subsection{Experiment results }

\begin{figure}[h]
    \centering
    \includegraphics[width=0.245\linewidth]{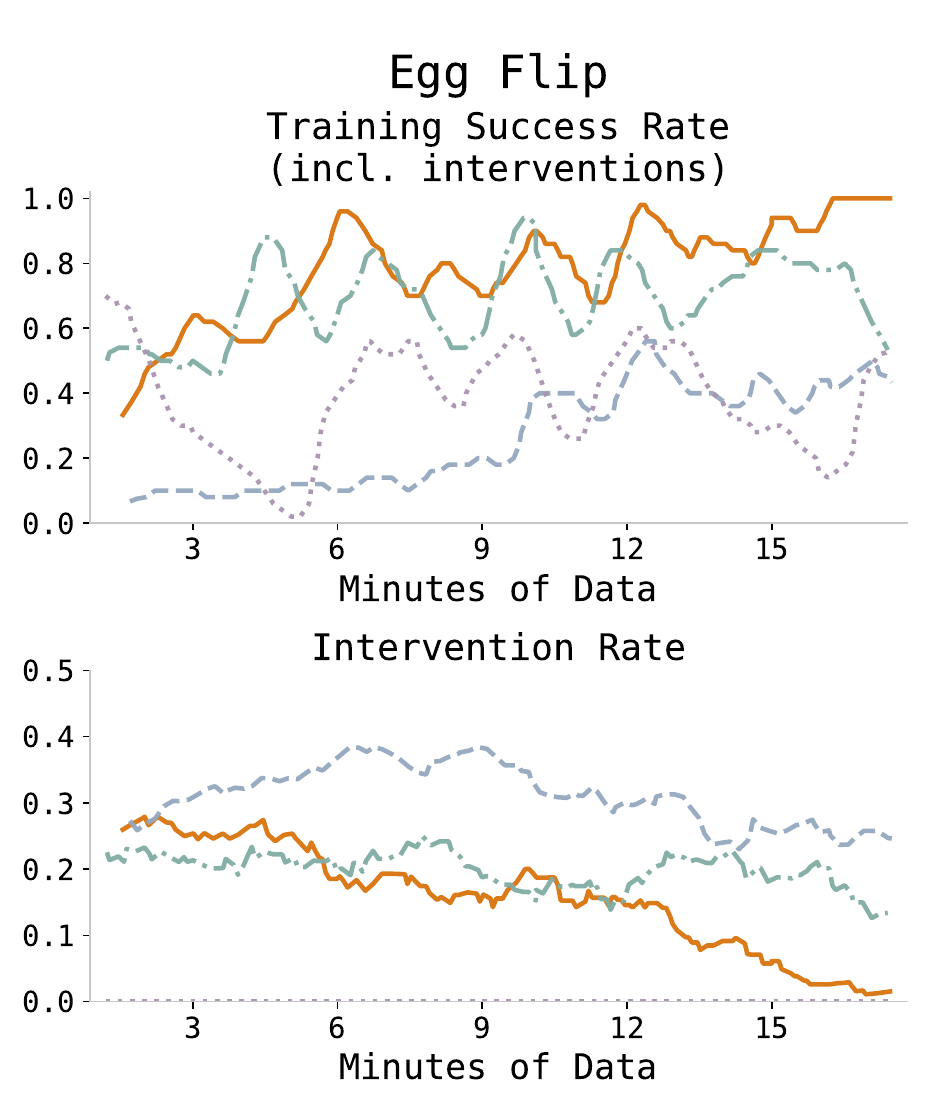}
    \hfill
    \includegraphics[width=0.245\linewidth]{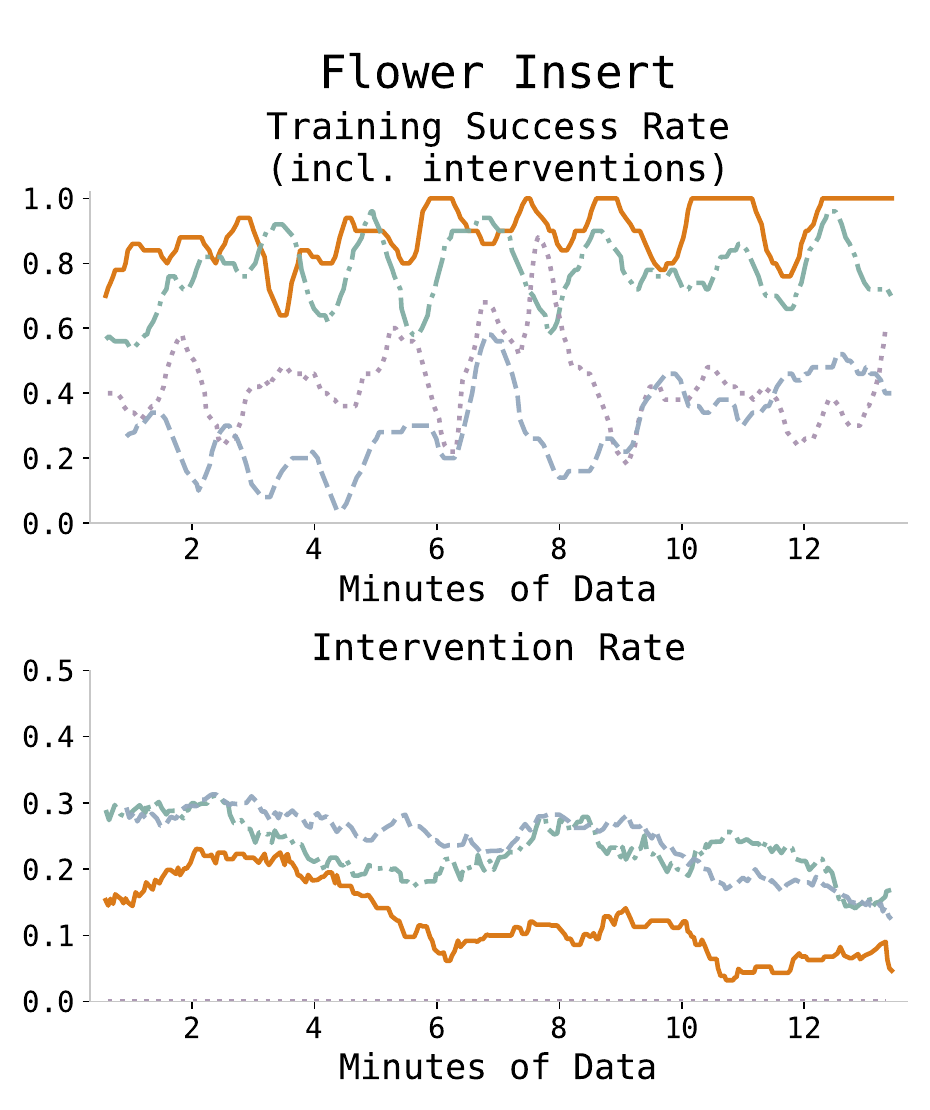}
    \hfill
    \includegraphics[width=0.245\linewidth]{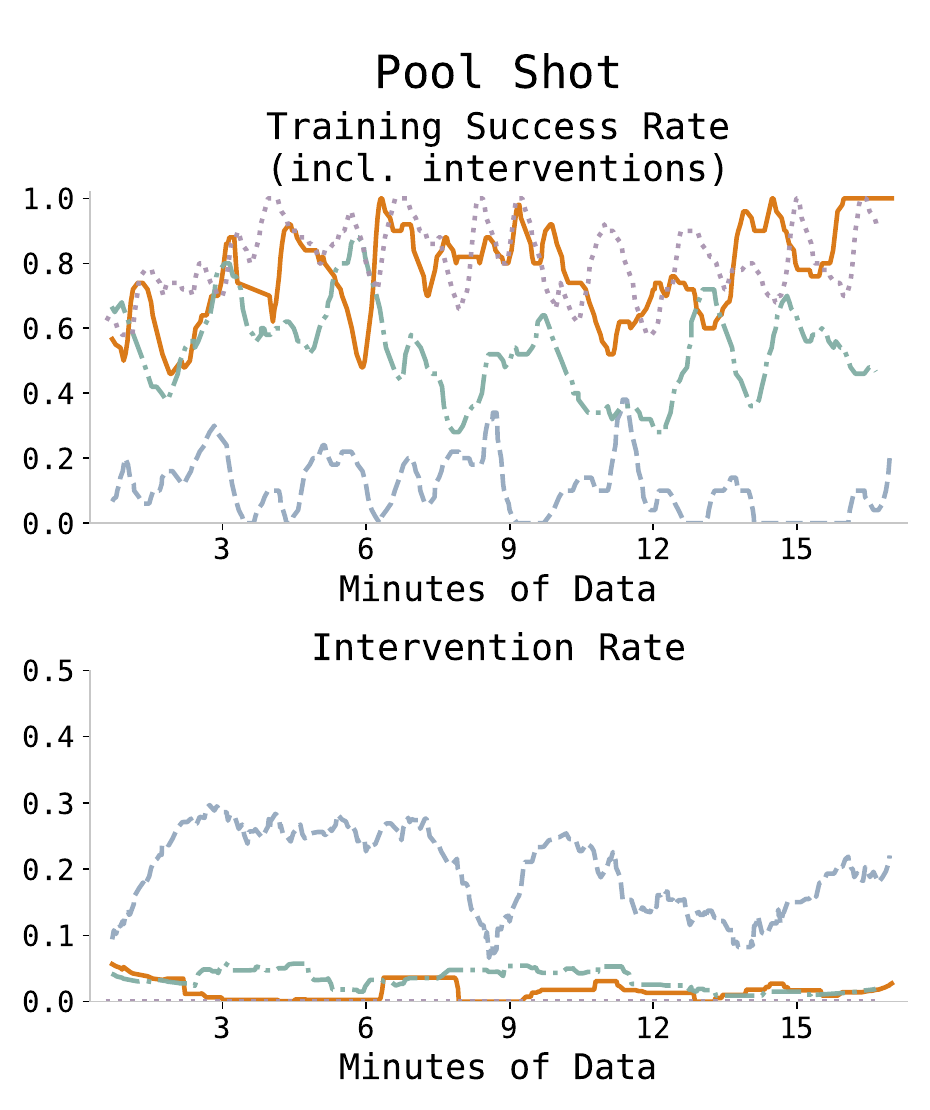}
    \hfill
    \includegraphics[width=0.245\linewidth]{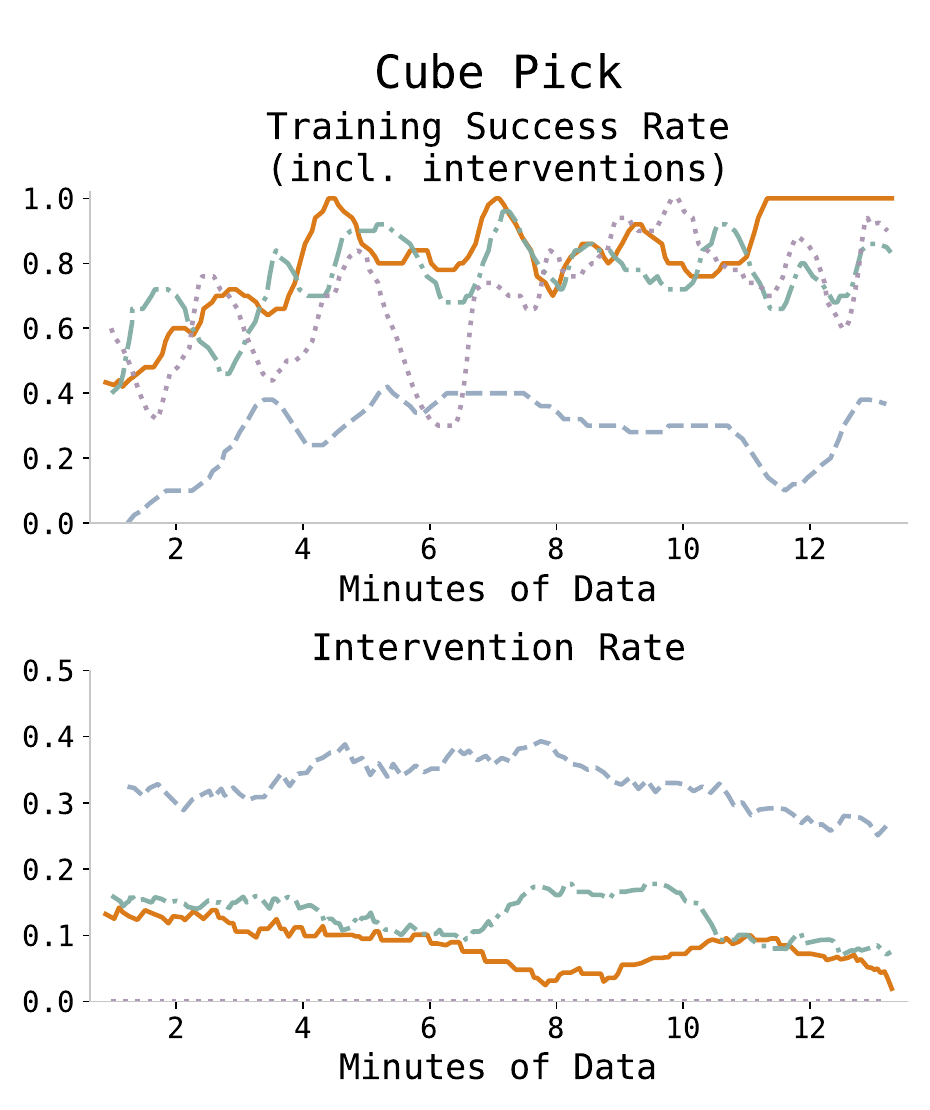}
    \hfill
    \\[0.3em]
    \includegraphics[width=0.245\linewidth]{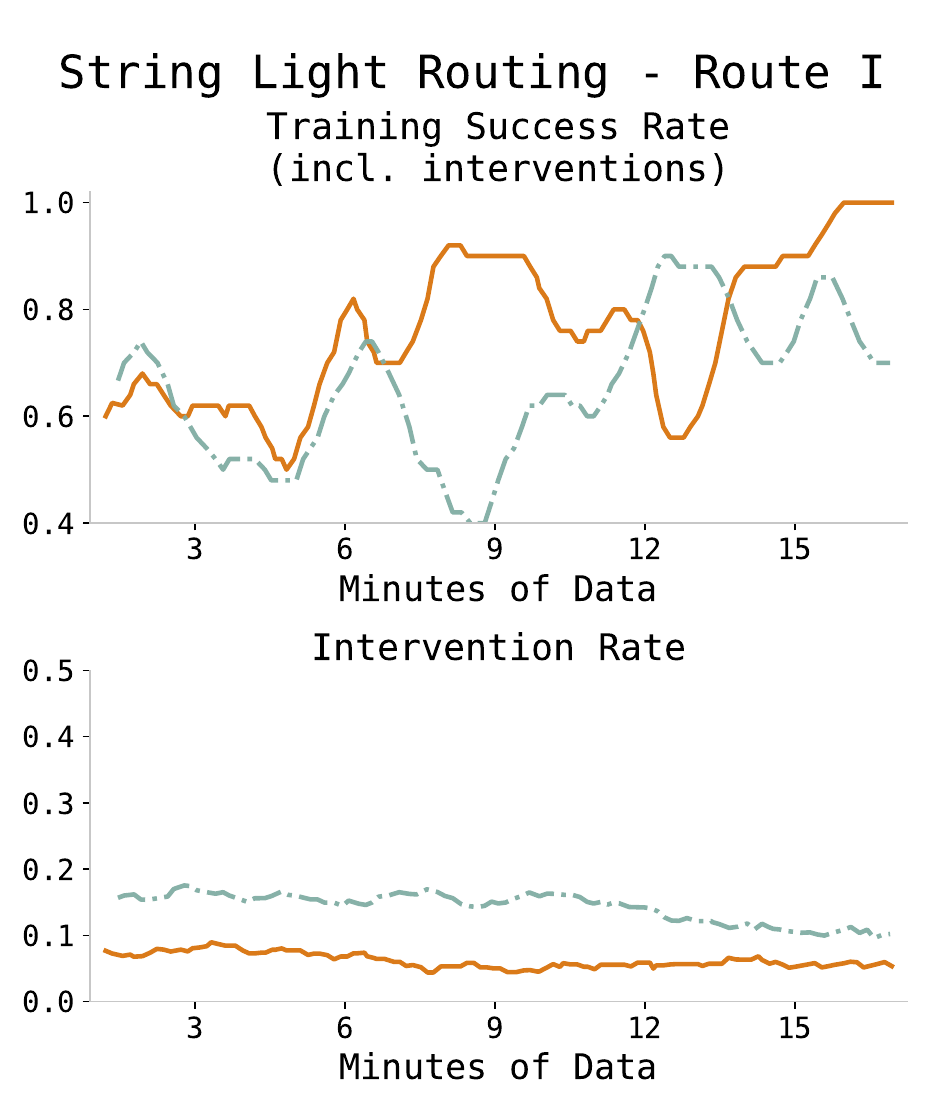}
    \includegraphics[width=0.245\linewidth]{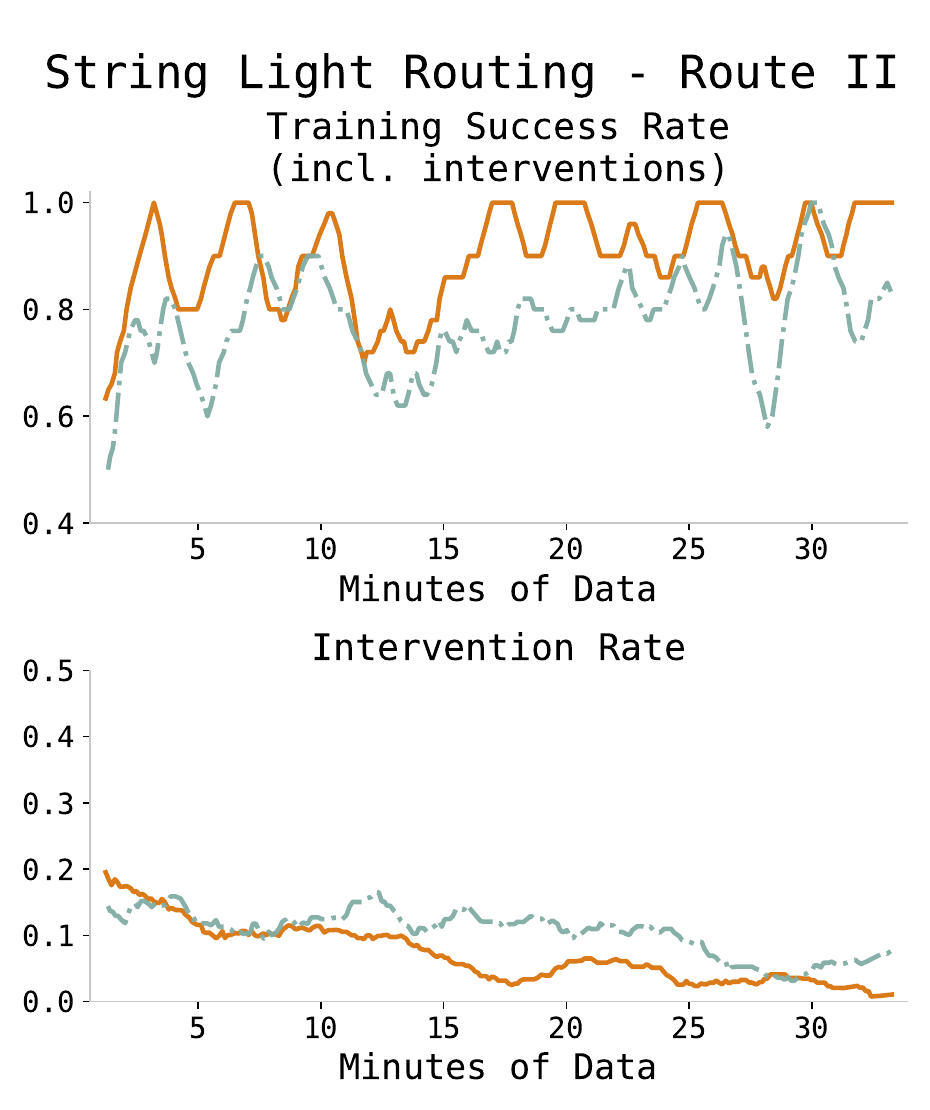}
    \hfill
    \includegraphics[width=0.245\linewidth]{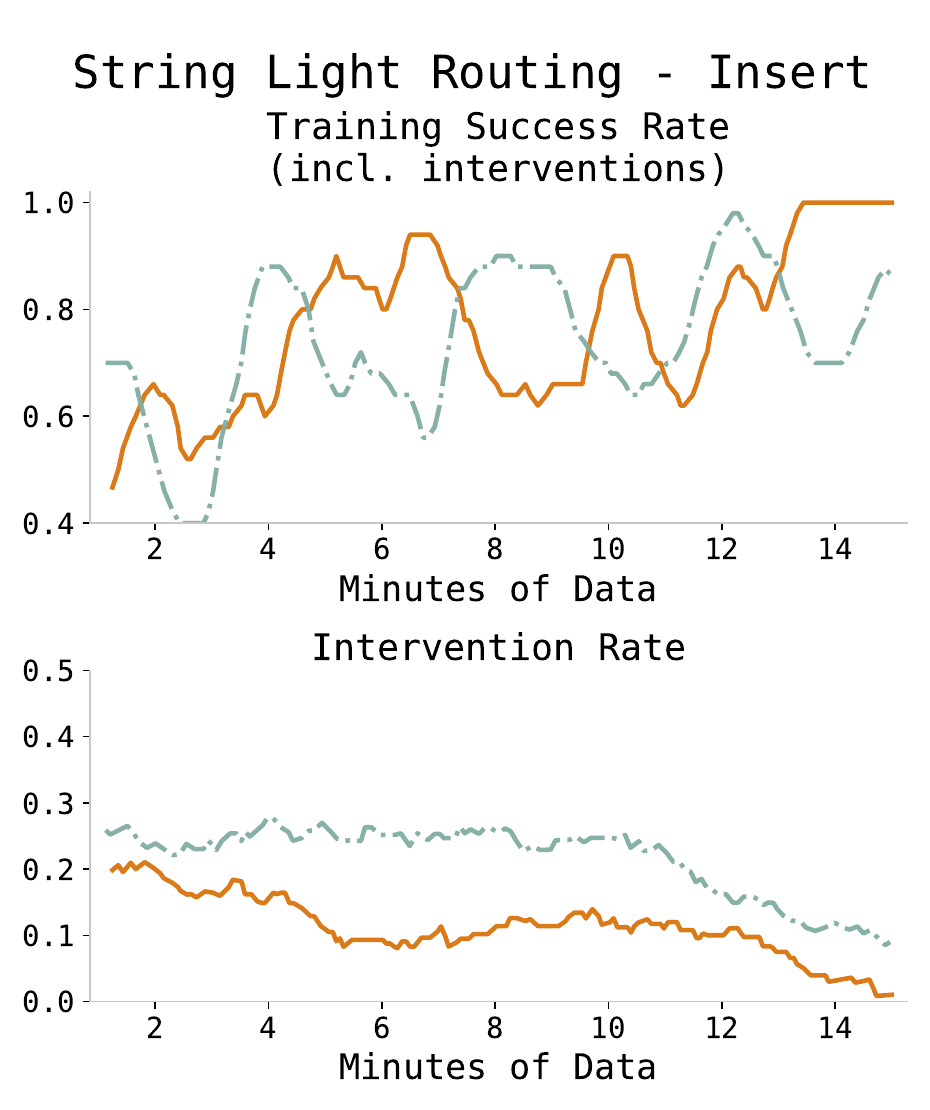}
    \hfill
    \includegraphics[width=0.245\linewidth]{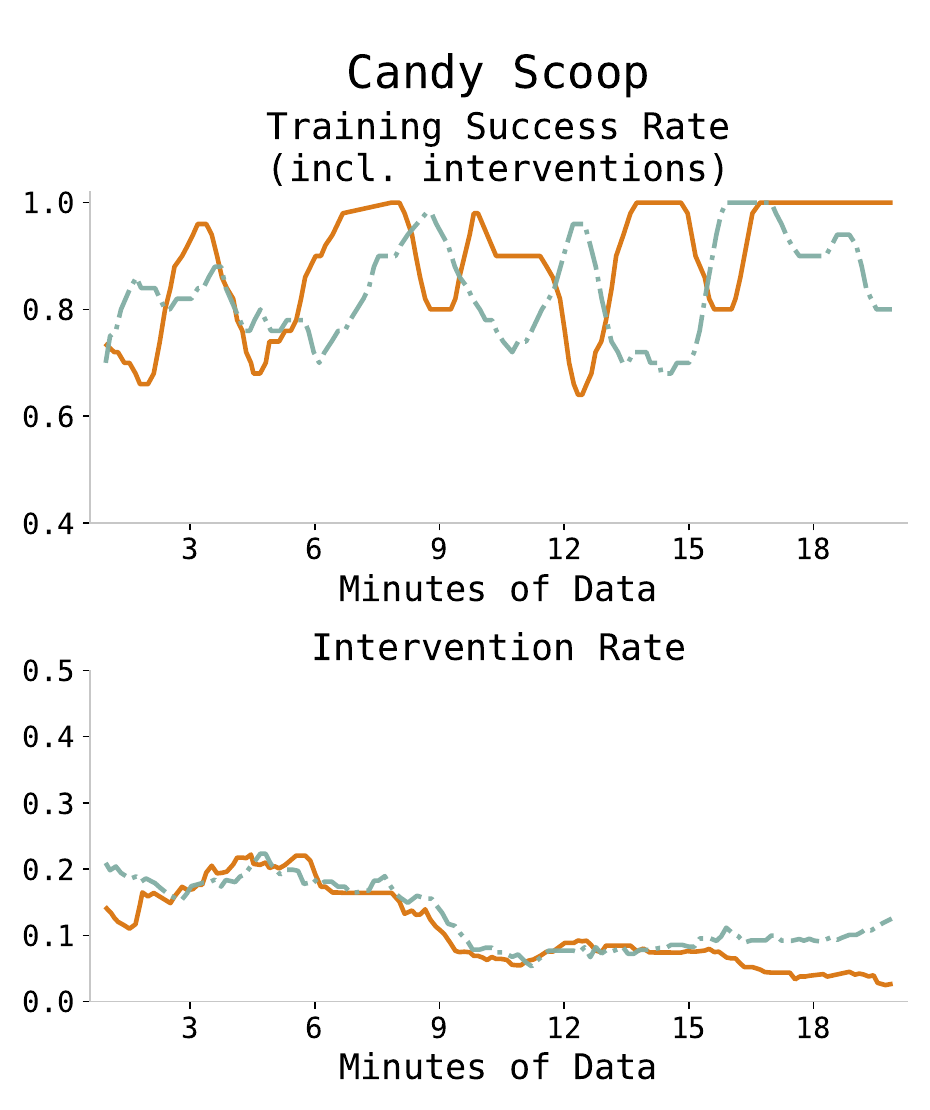}
    \\[0em]
    \includegraphics[width=0.6\linewidth]{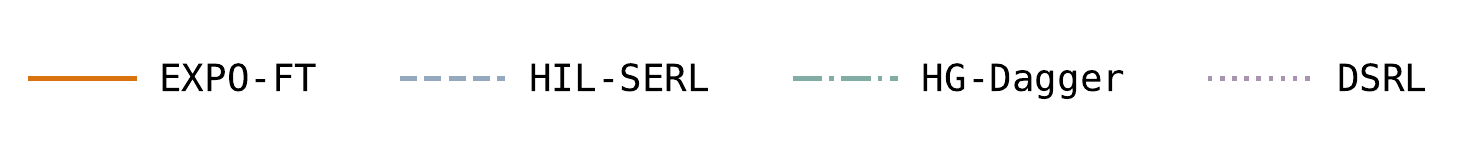}
    \caption{ \footnotesize \textbf{Training success and intervention rates across all tasks. Top row:} Egg Flip, Flower Insert, Pool Shot, Cube Pick. \textbf{Bottom row:} String Light Routing - Route I, String Light Routing - Route II, String Light Routing - Insert, Candy Scoop.}
    \label{fig:all_tasks_plots}
\end{figure}

\begin{table*}[t]
\centering
\setlength{\tabcolsep}{6pt}
\caption{ \footnotesize \textbf{Success rates across selected tasks comparing \ours{} to all points of comparison.} Performance is reported as successful trials out of 30. HIL-SERL (M) denotes HIL-SERL trained with additional samples, as the standard training budget is insufficient for it to begin learning on Cube Pick and Pool Shot. While HIL-SERL achieves highly reliable performance in its original evaluations, our experiments involve a substantially larger initial state space, under which HIL-SERL fails to achieve high performance consistently. }
\begin{tabular}{lcccccc}
\toprule
\multirow{2}{*}{Task} & \multicolumn{6}{c}{Success Rate (x/30)} \\
\cmidrule(lr){2-7}
 & SFT & HG-DAgger & DSRL & HIL-SERL & HIL-SERL (M) &  \ours{} \\
\midrule
Egg Flip                    & 16/30 & 18/30 & 15/30 & 13/30 & - & \textbf{30/30} \\
Cube Pick                   & 22/30 & 26/30 & 24/30 & 0/30 & 27/30 & \textbf{30/30} \\
Pool Shot                       & 23/30 & 14/30 & 25/30 & 1/30 & 13/30 &\textbf{30/30} \\
Flower Insertion                      & 14/30 & 24/30 & 12/30 & 8/30 & - &\textbf{30/30} \\
\midrule
Average                & 18.8/30 & 20.5/30 & 19/30 & 5.5/30 & 20/30 & \textbf{30/30} \\
\bottomrule
\end{tabular}
\label{tab:main_results_all}
\end{table*}

\begin{table*}[t]
\centering
\setlength{\tabcolsep}{8pt}

\caption{ \footnotesize \textbf{Comparison of success rates across all tasks} against SFT and HG-DAgger. Performance is reported as successful trials out of 30, with improvements over SFT shown in parentheses. }
\label{tab:main_results}
\begin{tabular}{lcccc}
\toprule
\multirow{2}{*}{Task} & \multirow{2}{*}{\shortstack{Online data  \\ (min)}} & \multicolumn{3}{c}{Success Rate (x/30)} \\
\cmidrule(lr){3-5}
 & & SFT & HG-DAgger & \ours{} \\
\midrule
Egg Flip                    & 18 & 16/30 & 18/30 & \textbf{30/30 (+88\%)} \\
String Light Routing - Route I   & 18 & 23/30 & 18/30 & \textbf{30/30 (+30\%)} \\
String Light Routing - Route II & 35 & 21/30 & 25/30 & \textbf{30/30 (+43\%)} \\
String Light Routing - Insert & 16 & 23/30 & 24/30 & \textbf{30/30 (+30\%)} \\
Candy Scoop                  & 20 & 22/30 & 28/30 & \textbf{30/30 (+36\%)} \\
Cube Pick                    & 14 & 22/30 & 26/30 & \textbf{30/30 (+36\%)} \\
Flower Insert                & 14 & 14/30 & 24/30 & \textbf{30/30 (+114\%)} \\
Pool Shot                       & 18 & 23/30 & 14/30 & \textbf{30/30 (+30\%)} \\
\midrule
Average                & 19.1 & 20.5/30 & 22.1/30 & \textbf{30/30 (+44\%)} \\
\bottomrule

\end{tabular}
\end{table*}

We now present the experimental results. The full results are shown in~\Cref{tab:main_results_all} and~\Cref{tab:main_results}. Across all eight tasks, \ours{} achieves an average success rate of 30/30, outperforming both HG-DAgger (22.1/30), SFT (20.5/30) on all tasks, and HIL-SERL (5.5/30), and DSRL (19/30) on the subset of tasks these methods were evaluated on. Comparing performance of the prior methods, the performance gap compared to HG-DAgger is most pronounced on tasks that require dynamic action or high precision — notably Egg Flip and Pool Shot — where imitation learning even with human corrections struggles to recover from compounding errors and learn precise actions, while \ours{} learns through self-improvement to correct and refine its behavior over training. Comparing to other RL methods, both HIL-SERL and DSRL fall significantly behind on tasks that involve a wider initial state distribution, for example Egg Flip, where the rich semantic representations encoded in the VLA initialization allow the policy to generalize across diverse initial states. Although DSRL leverages the pretrained VLA as its starting point, its optimization is constrained to modes already covered by the offline data, limiting its ability to acquire different behavior. HIL-SERL additionally struggles on the Pool Shot task in our setting as it requires highly precise motions to strike the ball into the hole, and the base RLPD algorithm simply struggles to learn that given the number of samples. Our approach, \ours{}, combines the strengths of these approaches: it initializes from a pretrained VLA prior, and then applies RL with online human-in-the-loop feedback to directly optimize performance.

The training data column in~\Cref{tab:main_results} reports the total interaction time required to reach the reported performance for all methods. \ours{} reaches reliable performance within 19.1 minutes of real-world interaction on average across tasks, demonstrating the high sample-efficiency of our RL finetuning system. This efficiency arises from both a strong prior provided by the VLA, which reduces the need for extensive exploration in the early stages of training, and a highly sample efficient finetuning algorithm and procedure. 

We also provide the training success rate curves and intervention rate during online training in~\Cref{fig:all_tasks_plots} and an additional average episode time in~\Cref{fig:all_tasks_plots_time} in Appendix~\Cref{appendix:time_plots}. As shown in the figures, \ours{} consistently improves throughout training while prior methods often struggle to reach a reliable performance.

Overall, our results demonstrate that \ours{} is both general and effective across tasks spanning a wide range of manipulation challenges, including dynamic interactions, precise object alignment, and deformable object handling. Using the same training approach across all tasks, \ours{} achieves the highest performance in every setting evaluated, validating the effectiveness and high sample efficiency of \ours{} for online RL finetuning of VLA models.

\section{Discussion} \label{sec:discussion}

In this work, we present \ours{}, a system for sample-efficient, reliable reinforcement learning finetuning of vision-language-action (VLA) models. Across a diverse set of challenging robotic manipulation tasks, \ours{} substantially improves upon the pretrained VLA baseline and achieves highly reliable performance within an average of 19.1 minutes of online robot interaction data — demonstrating that RL finetuning of VLAs is both practical and broadly beneficial. Despite the promising results, \ours{} has limitations. Our current pipeline requires human environment resets between episodes, which can introduce operational burden at scale. Automating the reset process remains an important direction for future work. Because we finetune a pretrained VLA model, which typically comprises billions of parameters, running training and inference at high frequencies remains computationally challenging. Addressing this constraint is another direction we leave for future work.

\section{Acknowledgments}

This work was in part supported by NSF CAREER, NSF \#1941722, RAI Institute, ONR grant N00014-22-1-2293, and ONR grant N00014-22-1-2621.


\clearpage


\nocite{dong2026tqlscalingqfunctionstransformers,dong2026valueflows}
\bibliography{example} 
\clearpage

\appendix

\section{Author Contributions} 

PD conceived the idea, devised the algorithm, and led the project. PD implemented the initial prototype, ran experiments and provided hands-on guidance on all experiments, analyzed and interpreted the results, and wrote and positioned the paper. KH maintained the research codebase, conducted a large number of experiments, and produced the figures and videos. TG conducted experiments and contributed to the design and development of the research codebase. DS and CF contributed to the research design, advised the project, and edited and positioned the paper.

\section{Additional Experiment Results}

\subsection{Training Episode Time} \label{appendix:time_plots}
In addition, we provide training episode time plots throughout online training. As shown in the plots, \ours{} consistently reduces the episode completion time as training progresses, indicating improved task efficiency and smoother policy execution during online adaptation.

\begin{figure}[h]
    \centering
    \includegraphics[width=0.245\linewidth]{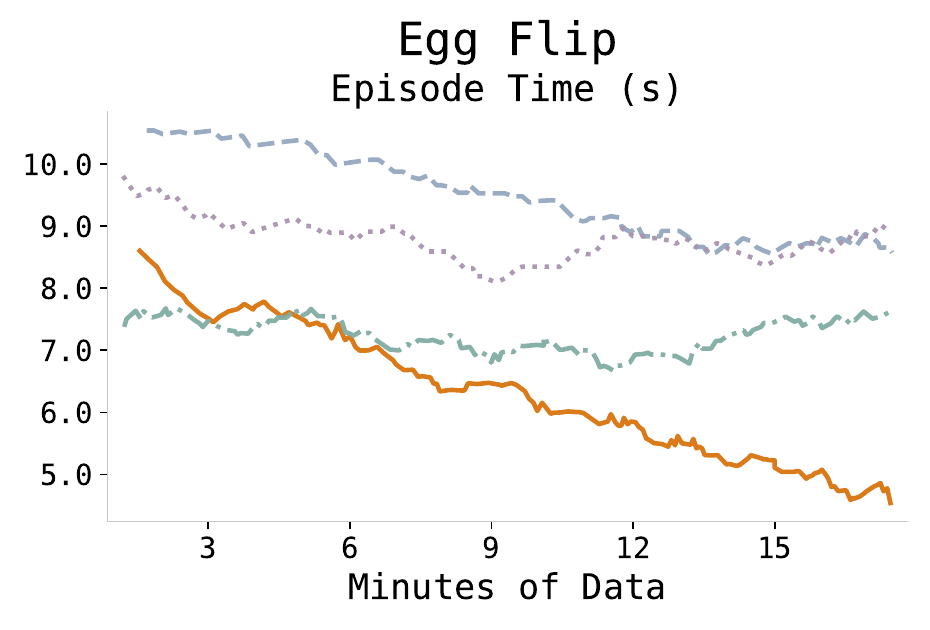}
    \hfill
    \includegraphics[width=0.245\linewidth]{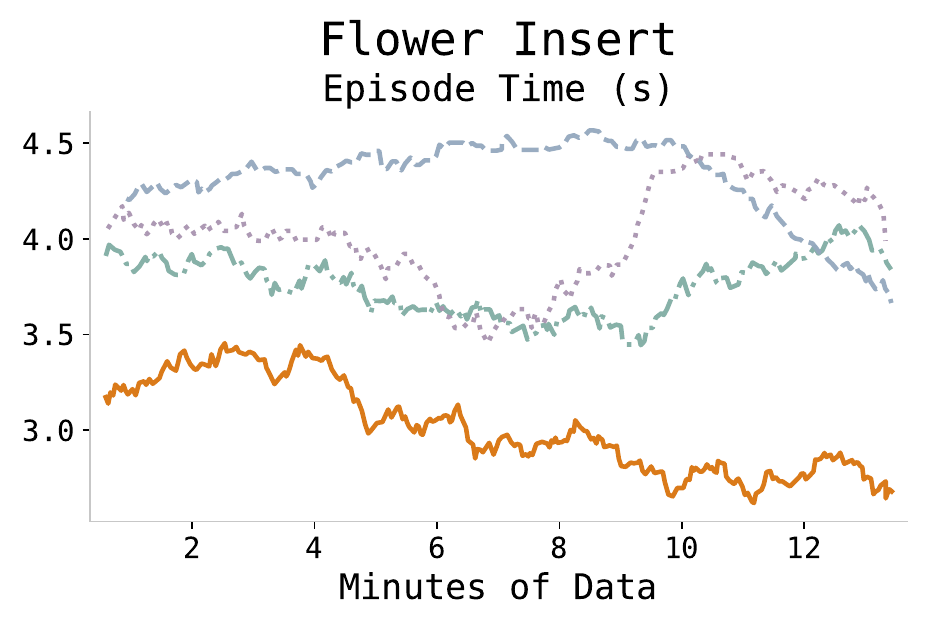}
    \hfill
    \includegraphics[width=0.245\linewidth]{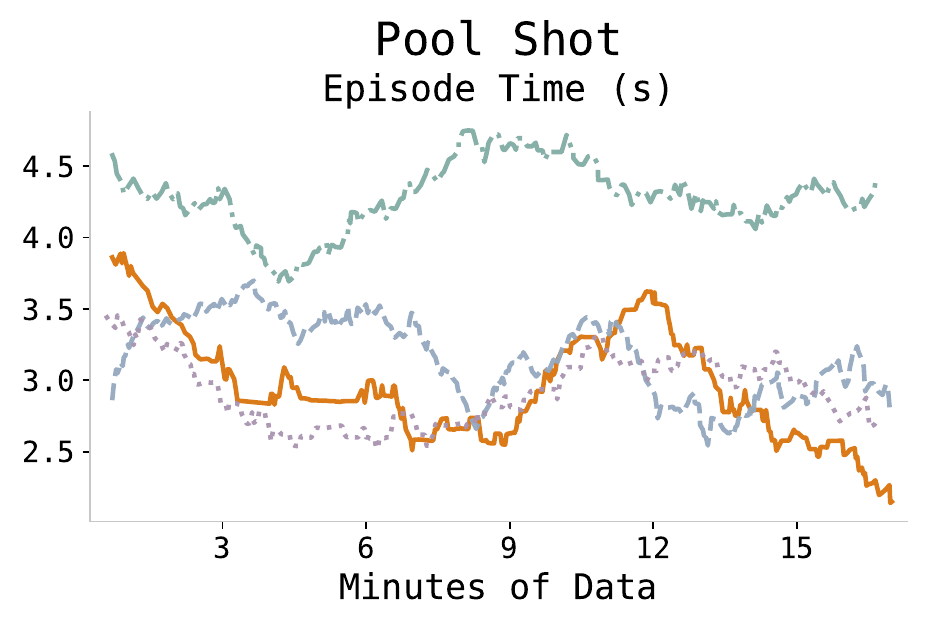}
    \hfill
    \includegraphics[width=0.245\linewidth]{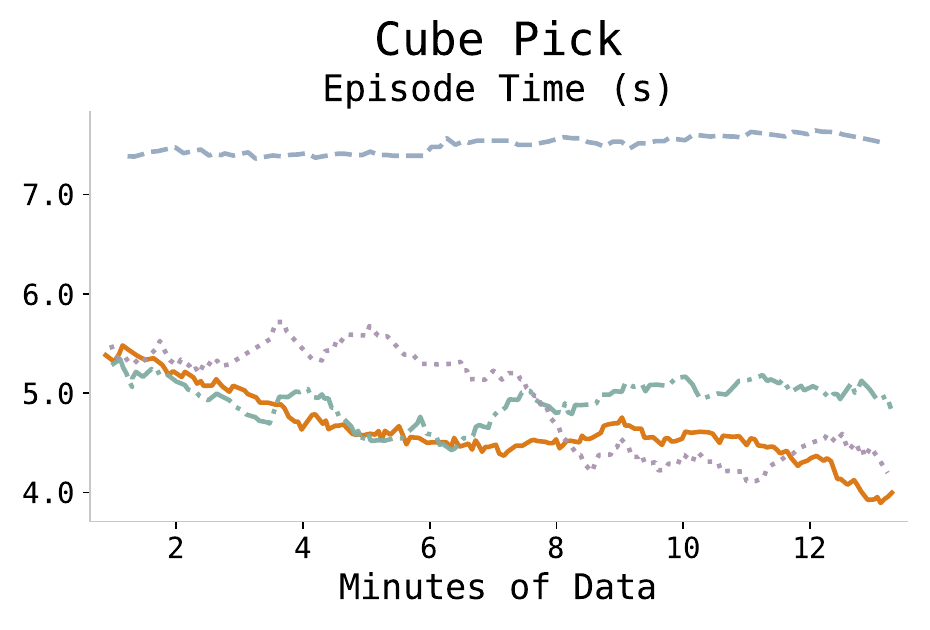}
    \hfill
    \\[0.3em]
    \includegraphics[width=0.245\linewidth]{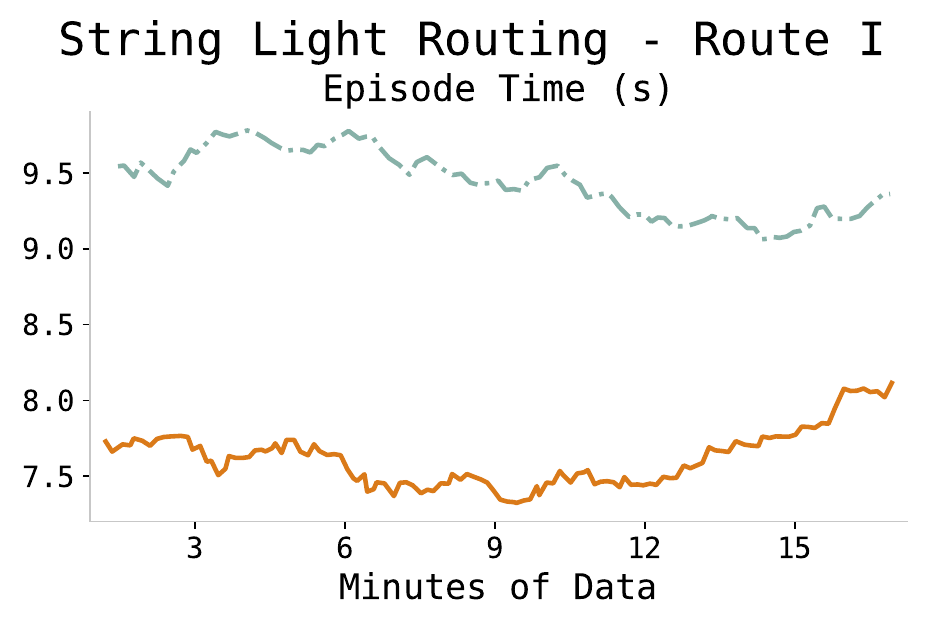}
    \includegraphics[width=0.245\linewidth]{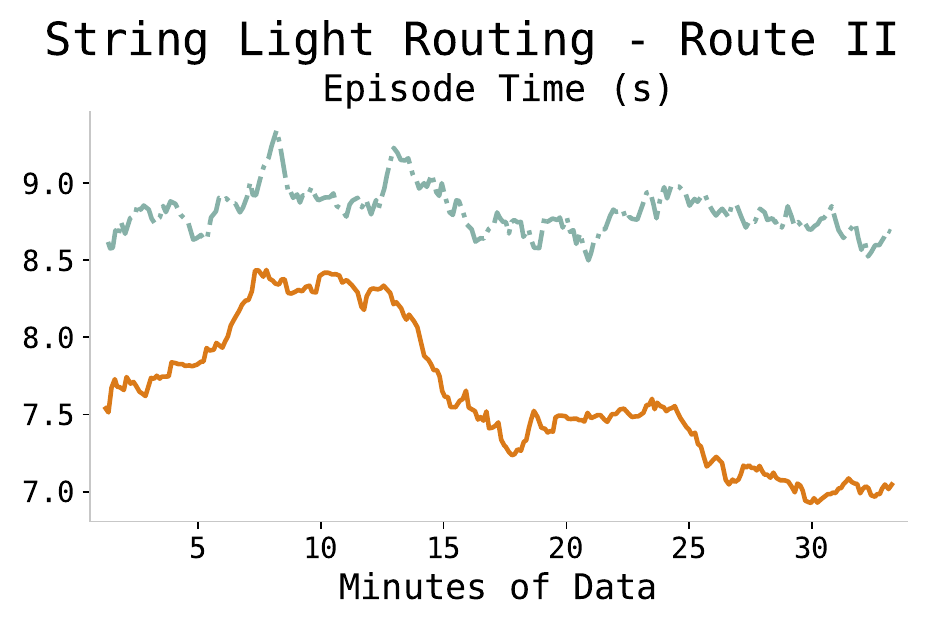}
    \hfill
    \includegraphics[width=0.245\linewidth]{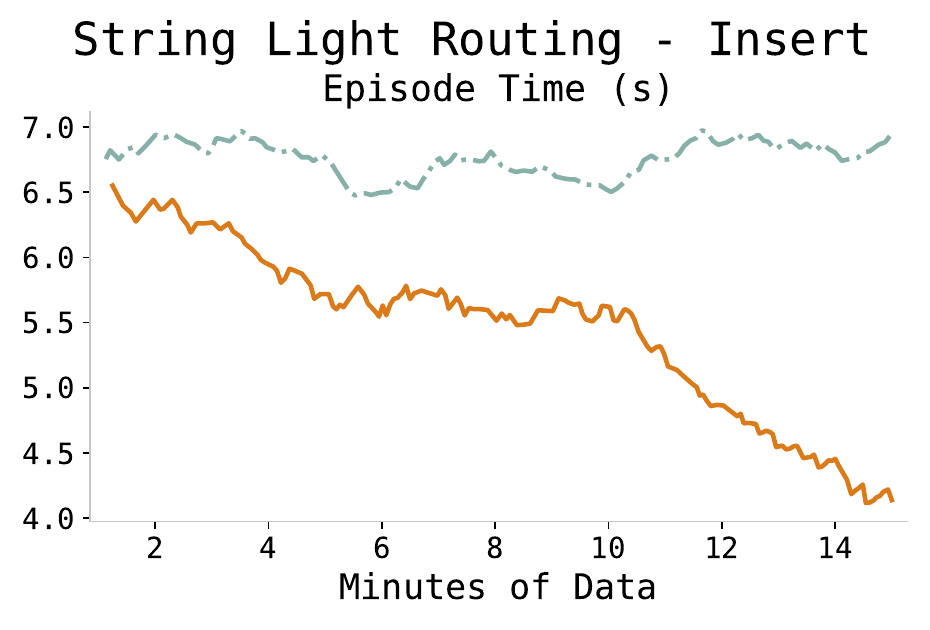}
    \hfill
    \includegraphics[width=0.245\linewidth]{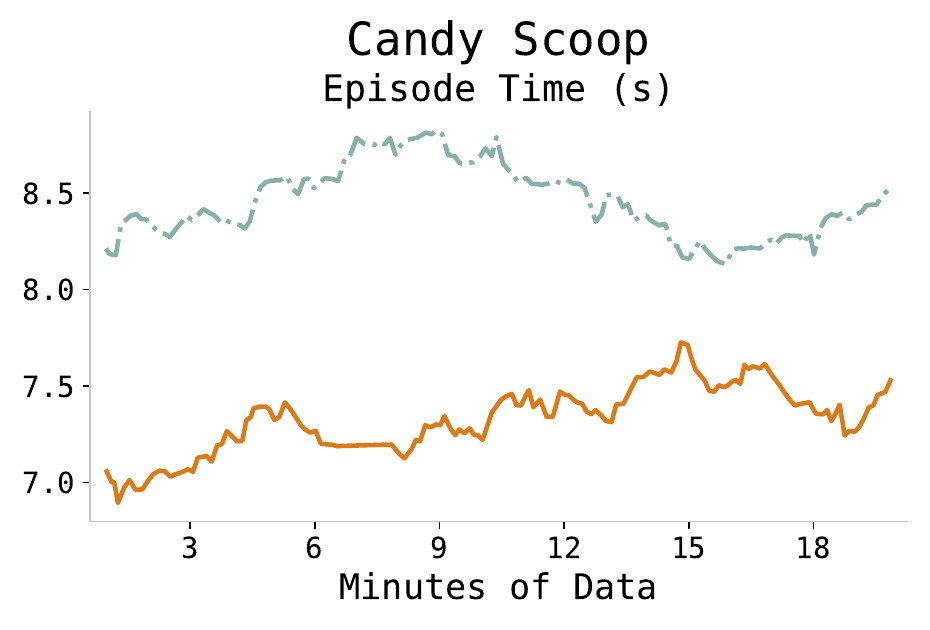}
    \\[0em]
    \includegraphics[width=0.6\linewidth]{figures/plots/legend_distinct_color.pdf}
    \caption{ \footnotesize \textbf{Episode Time across all tasks. Top row:} Egg Flip, Flower Insert, Pool Shot, Cube Pick. \textbf{Bottom row:} String Light Routing - Route I, String Light Routing - Route II, String Light Routing - Insert, Candy Scoop.}
    \label{fig:all_tasks_plots_time}
\end{figure}

\subsection{Which components are key for performance? }
\label{appendix:ablation}
\begin{figure}[h]
    \centering
    \includegraphics[width=1.0\linewidth]{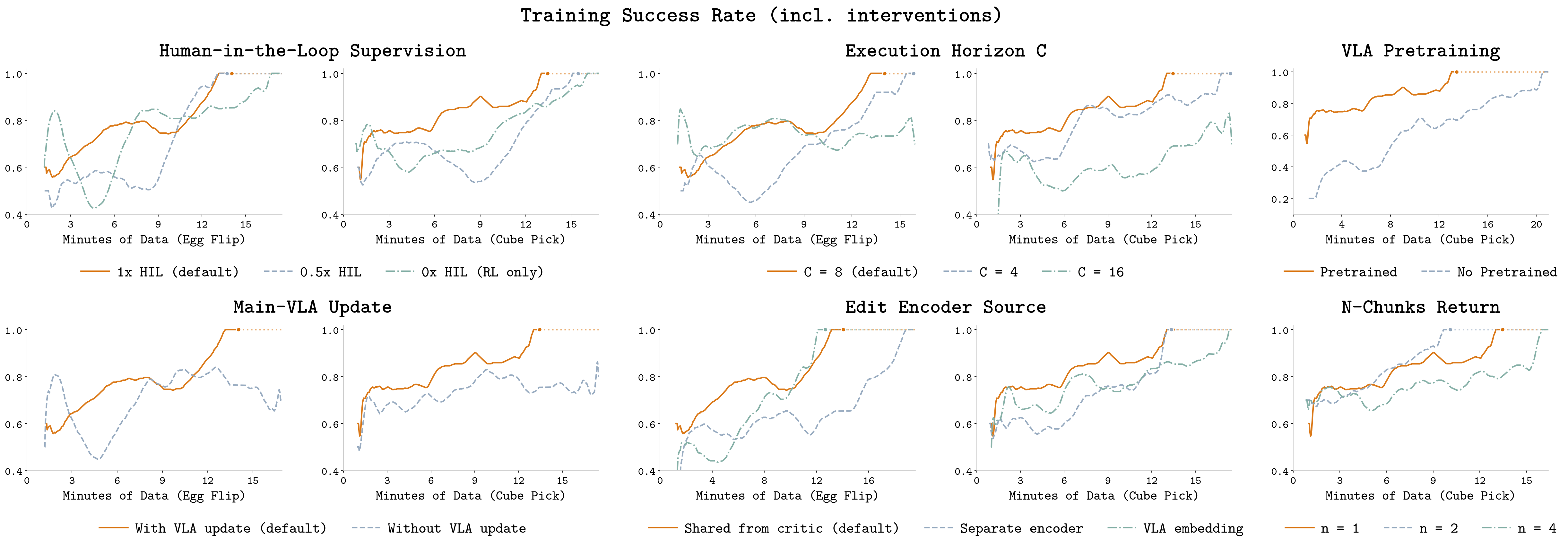}
    \caption{\footnotesize
    \textbf{Ablations of key components of \ours{}.}
    We evaluate the effects of human intervention, action chunk/replanning length, freezing the base VLA, the visual features used by the edit policy, initialization from a pretrained VLA, and the n-step returns. Experiments are conducted on Cube Pick and/or Egg Flip with 30 evaluation trials per setting.}
    \label{fig:ablation}
\end{figure}

We perform additional ablations to study which components of \ours{} are important for performance. We evaluate on Cube Pick and Egg Flip, with 30 evaluation trials per setting. The results are shown in \Cref{fig:ablation}. 

\paragraph{Human intervention.}
We isolate the contribution of human-in-the-loop (HIL) supervision by comparing \ours{} against (i) an RL-only variant that receives no interventions and (ii) a variant that receives half as many interventions ($0.5\times$ HIL). Both ablations eventually converge to $30/30$, but require substantially more online steps to do so. Thus, HIL is not necessary for the policy to improve, but it a way to improve sample efficiency, and RL alone still learns the task effectively. This suggests that the gains of \ours{} do not arise from additional human supervision, and that the intervention budget can be traded off against wall-clock interaction time depending on the availability of a human operator.

\paragraph{Action chunk and replanning length.}
We next vary the action chunk length used for execution. Without online RL fine-tuning, the base VLA obtains 19/30, 18/30, and 17/30 successes for chunk sizes of 4, 8, and 16, respectively. During online RL fine-tuning, replanning intervals of 4 and 8 achieve comparable final performance, whereas replanning every 16 steps performs worse. These results indicate that sufficiently frequent replanning can be important for closed-loop correction. We use a replanning interval of 8 in our main experiments as a compromise between policy reactivity and the additional inference cost associated with more frequent VLA queries.

\paragraph{Freezing the base VLA.}
By default, \ours{} jointly adapts the base VLA together with the edit policy. To isolate the importance of training the VLA, we freeze the base VLA and train only the remaining online components. In this setting, \ours{} does not reach 30/30 successes within the same interaction budget. This suggests that updating the VLA itself is important for efficient online fine-tuning. Nevertheless, freezing the VLA remains a reasonable lower-compute alternative when updating the full model is prohibitively expensive.

\paragraph{Visual features for the edit policy.}
We also ablate the visual representation provided to the edit policy. In our default configuration, the edit policy reuses visual features from the critic's ResNet-50 encoder, rather than using features from the VLA or learning a separate visual encoder. We compare all three choices. All evaluated feature variants eventually reach 30/30 successes; however, sharing the critic encoder leads to faster and more consistent convergence. We therefore use critic features by default, which additionally avoids introducing a separate visual backbone for the edit policy.

\paragraph{Initialization from a pretrained VLA.}
To determine whether a strong pretrained policy is necessary for \ours{}, we additionally train without initializing from a pretrained VLA. While the convergence is slower, \ours{} still reaches 30/30 successes in this setting. This suggests that pretraining accelerates online fine-tuning, but is not strictly necessary for it, and that \ours{} can be used to fine-tune a base policy to reliable performance even for randomly initialized VLA policies. 

\paragraph{N-step returns.}
Finally, we vary the N-step return used for critic learning. We observe meaningful differences in learning speed and performance across $n$-step return lengths. While we do not tune this for the main experiments, it can be another important component to affect performance.

Overall, identify which components of \ours{} matter most for performance and provide practical guidance on configurations that achieves the best sample efficiency.  

\section{Detailed Task Setting} \label{appendix:task_setting}

\begin{figure}[t]
  \centering
  \includegraphics[width=\linewidth]{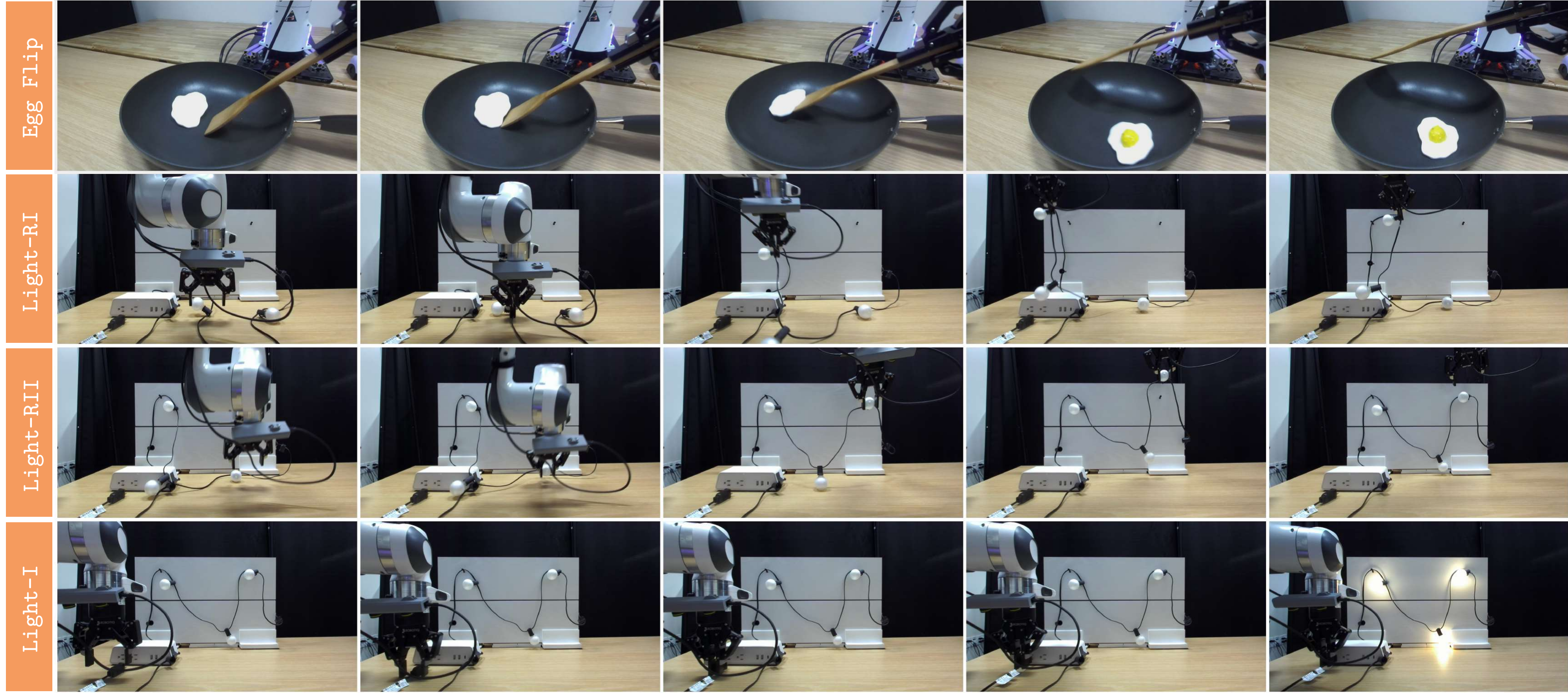}
  \vspace{0.5em}
  \includegraphics[width=\linewidth]{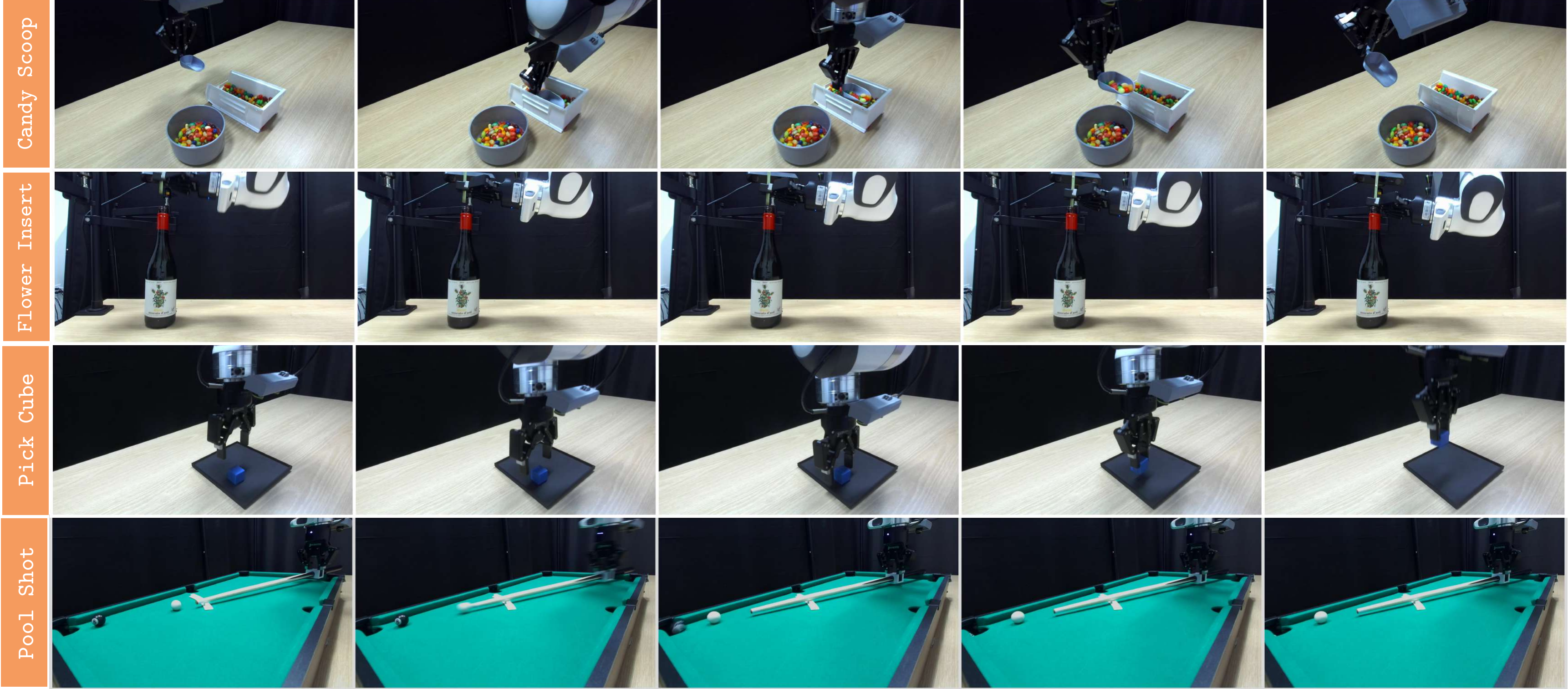}
  \caption{\textbf{Task strips demonstrating successful completion of each task. }}
  \label{fig:task_stips}
\end{figure}

\subsection{Task Setting Description}
Here, we provide detailed descriptions of the data collection process, reward specification, task success detector, reset mechanism and task randomization for each task. We also provide detailed task completion strips in~\Cref{fig:task_stips}.

\paragraph{Egg Flip.}
We pre-collect 25 demonstrations for this task. The task is counted as successful when the egg has been flipped to the opposite side while remaining inside the pan. To determine this, we record the egg's initial orientation and detect yolk pixels within a cropped region corresponding to the pan. The spatula remains grasped throughout training, and the robot returns to its initial position on each reset. A human reset is performed only if the egg falls outside the pan. The initial positions of the egg and spatula are randomized without constraints within approximately two thirds and one third of the pan, respectively.

\paragraph{String Light Routing - Route I.}
We pre-collect 40 demonstrations for this task. The task is counted as successful when the lightbulb's vertical position exceeds a fixed threshold and the gripper is open. The lightbulb position is tracked via pixel color detection. Human resets are performed between episodes. The lightbulb's initial position is randomized.

\paragraph{String Light Routing - Route II.}
We pre-collect 30 demonstrations for this task. The task is counted as successful when the lightbulb's vertical position exceeds a fixed threshold and the gripper is open. The lightbulb position is tracked via pixel color detection. Human resets are performed between episodes. The lightbulb's initial position is randomized.

\paragraph{String Light Routing - Insert.}
We pre-collect 25 demonstrations for this task. The task is counted as successful when the light turns on. During reset, the plug is dropped at a random location, and the robot automatically moves to a random reset position. The plug's initial position is randomized.

\paragraph{Candy Scoop.}
We pre-collect 20 demonstrations for this task. The reward classification for this task is split into two parts, both of which must succeed for the episode to be counted as successful. In the first part, we verify that candies are present in the scoop once the scoop is raised above a height threshold. In the second part, we check whether the candy count decreases to zero within a specified xyz region corresponding to the target container, indicating a successful pour (minor spillage is still considered successful). Candies are detected via pixel-based color matching. The scoop remains grasped throughout training, and the robot returns to its initial pose on each reset. A human reset is performed if there are too few candies remaining in the source container or if too many candies spill onto the table. The robot's initial position is randomized.

\paragraph{Cube Pick.}
We pre-collect 10 demonstrations for this task. The task is counted as successful when the gripper rises above a height threshold while performing a partial closing motion, indicating that an object is being grasped. During reset, the cube is moved to a random location and the robot moves to a random initial position. The cube's position is randomized across the entire black workspace plane.

\paragraph{Flower Insert.}
We pre-collect 25 demonstrations for this task. The task is counted as successful when the gripper is within a target region and no camera detects any portion of the flower stem below the mouth of the bottle. The stem is detected via pixel color matching. The gripper remains closed throughout training, and the robot moves to a random position on each reset. The robot's initial position is randomized.

\paragraph{Pool Shot.}
We pre-collect 30 demonstrations for this task. The task is counted as successful when three conditions are simultaneously satisfied: (1) the cue stick does not contact the black ball at any point during execution, (2) the black ball ends up in a pocket, and (3) the white ball does not end up in a pocket. Human resets are performed between episodes. The black ball's initial position is randomized along a fixed line.

\subsection{Task Initial State Randomization Space}

\begin{figure}[h]
  \centering
  \includegraphics[width=\linewidth]{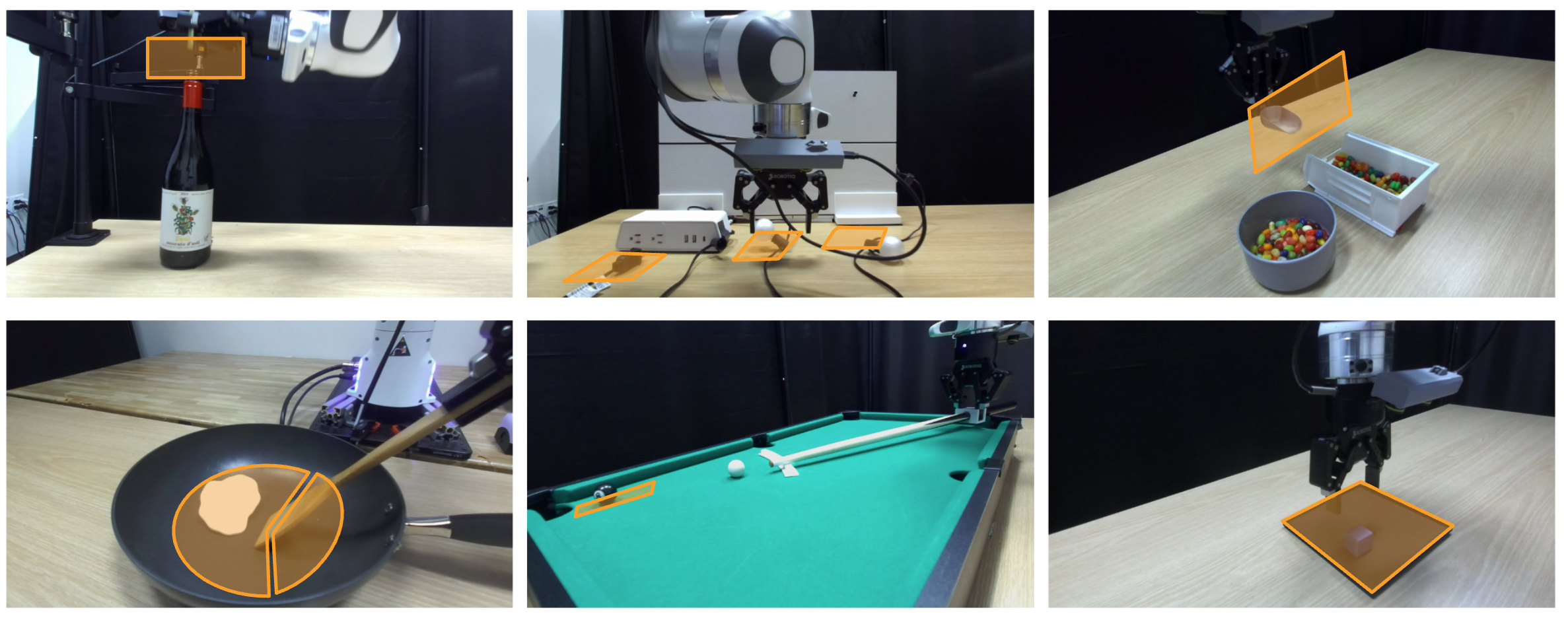}
  \caption{\textbf{Visualization of randomized initial state spaces for all tasks.} The orange regions indicate the randomized initialization areas used during training.}
  \label{fig:task_ini}
\end{figure}

Here, we provide visualizations of the randomized initial state space for each task, as shown in~\Cref{fig:task_ini}. The randomized regions are highlighted in orange. 

For Egg Flip, both the egg position and scoop position are randomized. For String Light Routing, the cable positions are randomized. For Candy Scoop, the gripper position is randomized. For Cube Pick, the gripper position is randomized. For Flower Insert, the gripper position is randomized. For Pool Shot, the position of the black ball is randomized.

\section{Detailed Training Setting} \label{appendix:train_setting}
\subsection{Model Structure/Training Detailed}
We instantiate \ours{} with $\pi0.5$~\cite{intelligence2025pi05visionlanguageactionmodelopenworld} as the base policy, initialized from a task-specific LoRA~\citep{hu2021loralowrankadaptationlarge} supervised-finetuning checkpoint and the matching normalization statistics for the robot setup. Input images are resized to 224×224 before encoding.

The value function is implemented as a RedQ-style~\citep{chen2021randomizedensembleddoubleqlearning} ensemble of 10 Q-networks with layer normalization. For each target value, we draw 2 networks at random from the ensemble, take their minimum to reduce overestimation, and apply slow Polyak averaging (standard soft update rate) toward a target ensemble. Discounting for multi-step action chunks is applied consistently with using 4 or 8 executed steps per replan (see Optimization).

Action sampling combines the base model with a small edit corrector. The base model draws 8 stochastic short-horizon action chunks per decision. edits are scaled by 0.05, 0.1 or 0.2 before being added to the base. Among the 16 candidates (8 base + 8 edit), the policy executes the chunk with the highest estimated Q-value (deterministic top‑Q selection, not a softmax).

The critic’s visual backbone is a ResNet-50~\citep{he2015deepresiduallearningimage} with stage depths (3, 4, 6, 3) and 64 filters per stage, producing a 512-dimensional image embedding; proprio is embedded to 64 dimensions and fused with the image representation for Q. Hidden MLP layers for the critic and edit modules use three layers of width 256.

Base-model image encoder weights are frozen during RL; critic image features are still learned. During updates we apply OpenPI-style~\cite{intelligence2025pi05visionlanguageactionmodelopenworld} augmentation to both camera views: a 95\% random crop with resize to $224{\times}224$, rotation in $[-5^\circ,5^\circ]$, and color jitter (brightness, contrast, saturation $\pm 0.1$). Each view and each of the current/next observations is augmented with an independent random draw.

\subsection{Optimization Hyperparameters}

In this section, we summarize the optimization hyperparameters used for \ours{}. We separate the hyperparameters into two groups: (1) global hyperparameters shared across all experiments, and (2) task-specific hyperparameters that are adjusted based on the manipulation difficulty and control requirements of each task.

All differentiable components are optimized using Adam with a learning rate of $3\times10^{-4}$. This includes the $Q$-ensemble, the tanh-bounded edit policy used to generate corrective actions, and the learnable temperature parameter in the maximum-entropy objective. The temperature parameter is initialized as $\alpha_0 = 1$.

We use a discount factor of $\gamma = 0.99$. The target $Q$-networks are updated using Polyak averaging with coefficient $\tau_Q = 5\times10^{-3}$. The target parameters of the base $\pi_{0.5}$ policy are updated using a faster Polyak averaging coefficient $\tau_\pi = 10^{-3}$.

Training uses replay minibatches of size $B = 64$. We employ a relatively high update-to-data ratio of $\mathrm{UTD}=20$, corresponding to $20$ gradient updates per environment-step-aligned batch of collected experience.

\begin{table}[h]
\centering
\caption{Shared optimization hyperparameters used across all experiments.} \vspace{4pt}
\label{tab:shared_hyperparameters}
\begin{tabular}{lc}
\toprule
Hyperparameter & Value \\
\midrule
Optimizer & Adam \\
Learning rate & $3\times10^{-4}$ \\
Discount factor $\gamma$ & $0.99$ \\
Initial temperature $\alpha_0$ & $1.0$ \\
Critic target update $\tau_Q$ & $5\times10^{-3}$ \\
Policy target update $\tau_\pi$ & $10^{-3}$ \\
Batch size & $64$ \\
Update-to-data ratio (UTD) & $20$ \\
\bottomrule
\end{tabular}
\end{table}

\Cref{tab:task_hyperparameters} summarizes the task-specific hyperparameters. The edit scale controls the magnitude of corrective edit actions added to the base policy. Tasks requiring precise manipulation generally use a smaller edit scale to avoid unstable corrections. 

The replanning interval $C$ determines how frequently the policy replans actions during execution. For tasks requiring higher precision or tighter feedback control, we use smaller replanning intervals to increase closed-loop responsiveness.

Finally, the number of updates per episode is selected such that the effective training ratio remains approximately close to one update per $40$ environment steps across tasks with different episode lengths.

\begin{table}[h]
\centering
\caption{Task-specific hyperparameters.}\vspace{4pt}
\label{tab:task_hyperparameters}

\resizebox{\linewidth}{!}{%
\begin{tabular}{lcccc}
\toprule
Task & Edit Scale & Replan Steps $C$ & Updates per Episode & Training Steps \\
\midrule
Egg Flip & $0.2$ & $8$ & $4$ & $10000$ \\
String Light Routing -- Route I & $0.05$ & $8$ & $4$ & $10000$ \\
String Light Routing -- Route II & $0.05$ & $8$ & $4$ & $20000$ \\
String Light Routing -- Insert & $0.05$ & $4$ & $3$ & $9000$ \\
Candy Scoop & $0.2$ & $8$ & $6$ & $12000$ \\
Cube Pick & $0.2$ & $8$ & $3$ & $8000$ \\
Flower Insert & $0.05$ & $4$ & $1$ & $8000$ \\
Pool Shot & $0.05$ & $8$ & $1$ & $10000$ \\
\bottomrule
\end{tabular}%
}
\end{table}

\subsection{Baseline implementations}

\textbf{HG-DAgger~\citep{kelly2019hgdaggerinteractiveimitationlearning}}
We train $\pi_{0.5}$~\cite{intelligence2025pi05visionlanguageactionmodelopenworld} with imitation learning only, where updates use supervised losses on human interventions together with offline data. Optimization uses Adam at $3\times10^{-4}$, a batch size of $64$, and the same update per episode as \ours{}. The per-episode update schedule and replan horizon $C$ are set identically to those of \ours{} on each task. We use \emph{synchronous} training (the learner runs in the main environment loop).

\textbf{HIL-SERL~\citep{luo2025precisedexterousroboticmanipulation}}
We follow the HIL-SERL~\citep{luo2025precisedexterousroboticmanipulation} training setup using the RLPD algorithm~\citep{ball2023efficientonlinereinforcementlearning} with the same $\pi_{0.5}$-style observation pipeline as our real-robot stack. The hyperparameters are: hidden sizes $(256,256,256)$, Adam with a learning rate of $3\times10^{-4}$, discount factor $\gamma=0.99$, batch size $64$, UTD ratio $10$ (we also experimented with a UTD ratio of $20$ as used in \ours{}, but found that UTD $10$ performs better under asynchronous training), initial temperature $0.1$, and Bellman backups \emph{without} an entropy term. The critic uses the same REDQ-style~\citep{chen2021randomizedensembleddoubleqlearning} ensemble as \ours{} (10 $Q$-networks, with 2 subsampled for the target). Training runs in an \emph{asynchronous} thread; the learner runs continuously and, in our setup, performs roughly \emph{one} full optimizer step per environment step on average.

\textbf{DSRL~\citep{wagenmaker2025steeringdiffusionpolicylatent}}
We use the DSRL built on frozen $\pi_{0.5}$ VLA. The hyperparameters are: hidden sizes $(256,256,256)$, Adam at $3\times10^{-4}$, $\gamma=0.99$, batch size $64$, UTD ratio $10$, initial temperature $0.1$, no entropy term in the backup, a REDQ-style critic (10 networks, with two used for the target), a $512$-D image latent and a $64$-D state latent with a ResNet ~\citep{he2015deepresiduallearningimage} $(3,4,6,3)$ of width $64$, state included in $Q$, and full image augmentation. Training is \emph{asynchronous}, with the same approximate effective rate of one update per environment step.

\end{document}